\documentclass{article}

\usepackage{authblk}
\usepackage[top=30truemm,bottom=30truemm,left=25truemm,right=25truemm]{geometry}
\usepackage[square,numbers]{natbib}

\usepackage{amsmath,amsfonts,bm}

\def\eqref#1{equation~\ref{#1}}

\def\1{\bm{1}}

\DeclareMathAlphabet{\mathsfit}{\encodingdefault}{\sfdefault}{m}{sl}
\SetMathAlphabet{\mathsfit}{bold}{\encodingdefault}{\sfdefault}{bx}{n}

\DeclareMathOperator*{\argmax}{arg\,max}
\DeclareMathOperator*{\argmin}{arg\,min}

\usepackage{hyperref}
\usepackage{url}

\usepackage{amsmath}
\usepackage{amssymb}
\usepackage{amsthm}
\usepackage{bm}
\usepackage{stmaryrd}

\usepackage{graphicx} 
\graphicspath{{./figure/}}

\usepackage[subrefformat=parens]{subcaption}

\usepackage{algorithm}
\usepackage[noend]{algpseudocode}
\algdef{SE}[DOWHILE]{Do}{DoWhile}{\algorithmicdo}[1]{\algorithmicwhile\ #1}%

\usepackage{booktabs}

\usepackage{hyperref}
\usepackage{url}

\usepackage{multirow}

\theoremstyle{definition}
\newtheorem{dfn}{Definition}

\newcommand{\acceptedversionnote}{
    \textit{This is the author’s accepted manuscript. The final version is published in Neural Networks, Elsevier, and is available via\\
    \texttt{https://doi.org/10.1016/j.neunet.2025.107610}}
}

\title{Minimal Sufficient Views: A DNN model making predictions with more evidence has higher accuracy}
\author[1]{Keisuke Kawano$^\ast$}
\author[1]{Takuro Kutsuna\thanks{Contributed equally.\\
\acceptedversionnote}}
\author[2]{Keisuke Sano}
\affil[1]{\normalsize Toyota Central R\&D Labs., Inc.}
\affil[2]{\normalsize DENSO CORPORATION}
\date{}

\begin{document}
\maketitle
\begin{abstract}
    Deep neural networks (DNNs) exhibit high performance in image recognition; however, the reasons for their strong generalization abilities remain unclear. A plausible hypothesis is that DNNs achieve robust and accurate predictions by identifying multiple pieces of evidence from images. Thus, to test this hypothesis, this study proposed minimal sufficient views (MSVs). MSVs is defined as a set of minimal regions within an input image that are sufficient to preserve the prediction of DNNs, thus representing the evidence discovered by the DNN. We empirically demonstrated a strong correlation between the number of MSVs (i.e., the number of pieces of evidence) and the generalization performance of the DNN models.
    Remarkably, this correlation was found to hold within a single DNN as well as between different DNNs, including convolutional and transformer models. This suggested that a DNN model that makes its prediction based on more evidence has a higher generalization performance.
    We proposed a metric based on MSVs for DNN model selection that did not require label information. Consequently, we empirically showed that the proposed metric was less dependent on the degree of overfitting, rendering it a more reliable indicator of model performance than existing metrics, such as average confidence.
\end{abstract}

\section{Introduction} \label{sec:introduction}
Deep neural networks (DNNs) perform well in various fields, including computer vision.
Although several approaches have been proposed to explain the generalization ability of DNNs theoretically or empirically~\citep{jacot2018neural,shrikumar2017learning,NIPS2017_10ce03a1,sturmfels2020visualizing,zhang2021understanding}, the reasons for the high generalization ability of DNNs remain unclear.

When considering recognition tasks in the vision domain, humans naturally recognize an object class in an image from multiple perspectives.
For example, we can identify several pieces of evidence in an image to distinguish cats from dogs: the shape of their eyes, radar-dish-shaped ears covered with fur, or striped patterns on their bodies.
It is plausible that the accuracy of human recognition of an object in an image of a cat increases as he/she finds more evidence of a cat.
Thus, the question addressed in this study is whether such a relationship holds true for image recognition using DNN models.

To address our research question, we proposed the concept of \emph{minimal sufficient views (MSVs)} for mathematically defining multiple pieces of evidence for the predictions made by a DNN model.
Considering a DNN model and input image, MSVs were defined as a set of minimal and distinct regions in the image, each of which was sufficient to preserve the model's prediction of the image.
To enable the efficient computation of MSVs for real data, we proposed a simple yet effective algorithm for estimating MSVs wherein the minimality condition was relaxed to balance the computational burden and quality of the estimated MSVs.
For example, the proposed algorithm can compute meaningful MSVs for prediction by ResNet-101~\citep{he2016deep} on an image from the ImageNet~\citep{deng2009imagenet} validation set in less than one second on average.

\begin{figure}[tbp]
    \centering
    \includegraphics[width=1.0\linewidth,clip]{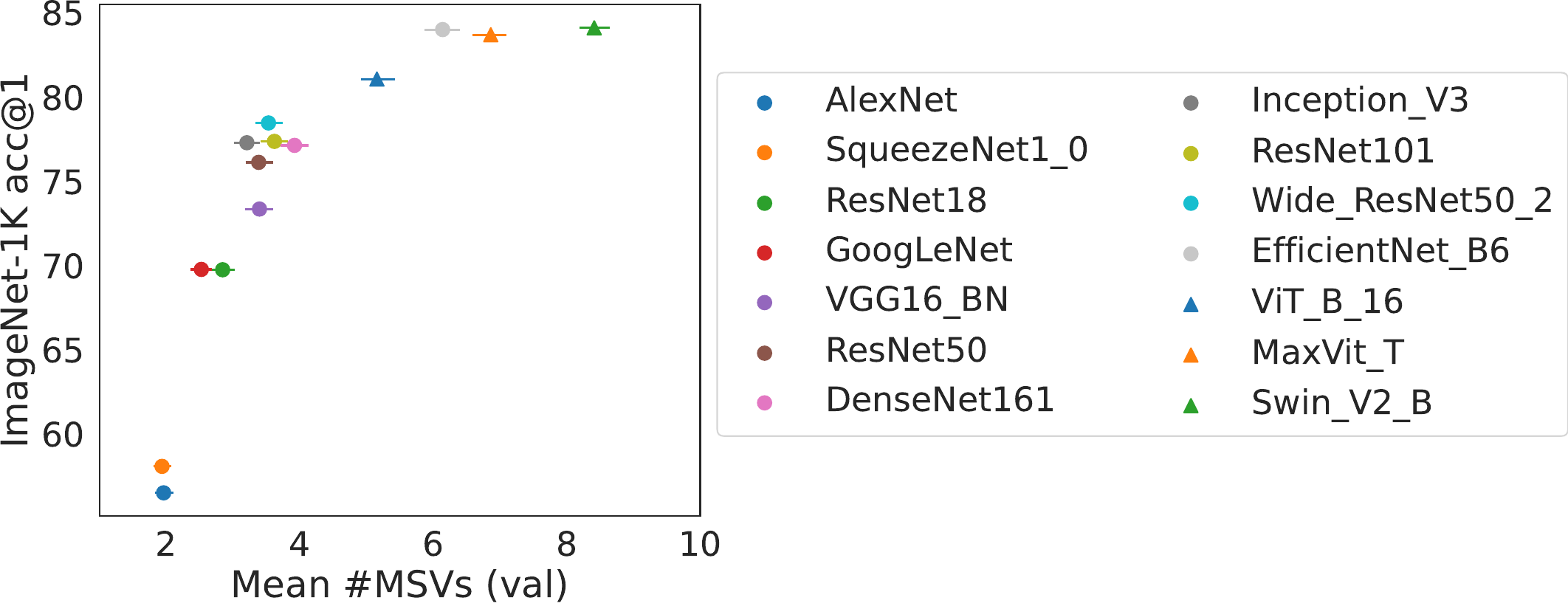}
    \caption{Average number of estimated MSVs for randomly sampled 1000 images from the ImageNet validation set (x-axis) and prediction accuracy on all data in the ImageNet validation set (y-axis). No label information is used to compute the MSVs. ImageNet-trained DNN models obtained from TorchVision are used in the evaluation (listed in the legend). See Section~\ref{sec:experiments} for the experimental details.}
    \label{fig:val_acc_msv}
\end{figure}

Our experimental results indicated that the number of estimated MSVs varied depending on the DNN model and input image. Moreover, a clear relationship was observed between the number of estimated MSVs and the accuracy of the prediction for the image.
This result supported our hypothesis that the more evidence a prediction has, the more accurate it is.
Remarkably, such a relationship was found to hold within a single DNN as well as between different DNNs, including convolutional and transformer models. This is shown in Figure~\ref{fig:val_acc_msv}, where we evaluated the average number of MSVs of each model (x-axis) as a metric to predict the generalization performance of the model (y-axis).
Our results suggested that the amount of evidence used in a prediction may be an important factor in understanding the generalizability of DNNs.

From a practical perspective, Figure~\ref{fig:val_acc_msv} implies that we can select the best performing DNN model using an unlabeled dataset because label information is \emph{not} required to compute MSVs.
Our experiments also showed that our metric was less dependent on the degree of overfitting of the model, that is, the score computed using the training dataset itself was close to the score computed using the holdout validation set. This was in contrast to the case with existing metrics such as the average confidence~\citep{9710388}.
This property is particularly useful because it eliminates the need to prepare holdout data when training and selecting the models.

The contributions of this study are summarized as follows:
\begin{itemize}
    \item We defined MSVs as a set of evidence for the prediction by a DNN model on a particular image (Section~\ref{sec:msv}).
    \item We proposed an efficient algorithm for estimating MSVs on real data (Section~\ref{sec:greedy}).
    \item We empirically showed that the number of estimated MSVs was related to prediction accuracy, and a prediction model with a larger average number of MSVs achieved higher generalization performance (Section~\ref{sec:experiments}).
\end{itemize}
The remainder of this paper is organized as follows.
Section~\ref{sec:related_work} reviews the existing work related to MSVs. Section~\ref{sec:proposed_method} defines MSVs and proposes a greedy algorithm to estimate MSVs.
We present the empirical results in Section~\ref{sec:experiments}. Finally, Section~\ref{sec:conclusion} summarizes this study.

\section{Related work} \label{sec:related_work}
Certain studies have attempted to explain the generalizability of DNNs from the perspective of the multiple features used in the prediction.
\citet{allen2020towards} introduced the concept of \emph{multi-view} for classification problems to analyze the generalization performance of DNNs, especially in ensemble or distillation settings.
In the multiview concept, it is assumed that there are multiple features per class, and each input comprises a combination of features.
For example, the ``car'' class is assumed to have several features, such as headlights, windows, and wheels, and each ``car'' image consists of a certain of these features.
\citet{allen2020towards} theoretically showed that the performance of DNN models can be improved by considering ensembles or distillation when the data exist in the multi-view structure.
However, the relationship between the multi-view hypothesis and the generalizability of standard (non-ensemble or non-distilled) DNN models was not discussed in~\citep{allen2020towards}.
In addition, \citet{allen2020towards} did not address a method for estimating the multiview used in the prediction of a particular input and model.

DNN model selection, including the model structure and training hyperparameters, is typically performed using a holdout-labeled validation dataset.
However, it is useful to use an unlabeled dataset for model selection, because labeling costs are generally high.
A few studies have proposed metrics for DNN model selection that can be evaluated using an unlabeled dataset.
In a related study, \citet{9710388} proposed the \emph{difference of confidence (DoC)}, which is the difference between the average confidence of the predictions for the training and test sets, as a measure to predict the performance degradation of DNN models under a distribution shift.
It was also reported in \citep{9710388} that the simple average confidence is a good predictor of the prediction accuracy of DNNs.
Because we did not consider the distribution shift situation in this study, we included the average confidence as a baseline in our experiment for predicting DNN performance without access to labeled data.

Extensive research has been conducted on explainable AI (XAI).
A primary approach in XAI research is to interpret the prediction results of a DNN model for a particular input by showing important parts of the image for prediction~\citep{selvaraju2017grad,ribeiro2016should,sundararajan2017axiomatic,shrikumar2017learning,fong2017interpretable,Petsiuk2018rise}.
In this sense, the proposal of MSVs is a type of XAI method because they estimate important parts of a given image from the viewpoint of preserving the prediction of a DNN for the image.
Many existing XAI methods in this category display their analysis results in a single-color heatmap, wherein the intensity of the color indicates the degree of importance.
By contrast, the proposed MSVs estimate a set of partial features in the image, each of which is sufficient to maintain the prediction result and can only be properly expressed with multiple colors. This results in a more thorough understanding of the prediction of complex nonlinear models, such as DNNs.

Several approaches have attempted to explain DNN prediction by examining the concepts used in the prediction~\citep{TCAV,IBD}, wherein the prediction is interpreted through the lens of the estimated concepts.
However, these approaches typically require additional information about concept candidates. To the best of our knowledge, no research in this area has investigated the relationship between prediction performance and the number of concepts used in the prediction.

\citet{carter2019made} proposed \emph{sufficient input subsets (SIS)} to interpret DNN prediction, where a set of multiple features similar to MSVs was considered.
\citet{carter2021overinterpretation} proposed \emph{Batched Gradient SIS (BG-SIS)} as an extension of SIS to improve computational efficiency and enable application to images of larger size, as in the ImageNet dataset.
However, neither of these studies discussed the relationship between the number of estimated SIS and the prediction accuracy of DNNs.
We empirically compared the proposed MSVs with SIS and showed that MSVs exhibited a clearer relationship with the generalizability of DNNs.
From an algorithmic perspective, BG-SIS requires gradient computation to accelerate the SIS search. However, the proposed algorithm did not use gradients to compute MSVs, thus allowing its application to black-box models, where gradients are difficult to compute.
Our experiments showed that the proposed algorithm could compute MSVs in less than a second on average for an image from the ImageNet dataset. This is in contrast to BG-SIS, which requires more than a minute on average to compute SIS.

Counterfactuals constitute another line of research on XAI~\citep{wachter2017counterfactual,mothilal2020explaining}.
Given a model and an input, these approaches attempt to find (or generate) another input, called the counterfactual, that is as close as possible to the original input, whereas the predicted outcome is different from the original.
Subsequently, the prediction is interpreted by showing the difference between the original input and its counterfactual.
By contrast, the algorithm proposed in this study attempted to find the smallest part of the input that preserved the prediction.

\section{Proposed method} \label{sec:proposed_method}
In this section, we formally define MSVs and propose a greedy algorithm to estimate them.
Before moving on to formal definitions, we first present a specific example to help with understanding.

\subsection{Example of estimated MSVs} \label{sec:msv_example}
\begin{figure}[tb]
    \centering
    \includegraphics[width=1.0\linewidth,clip]{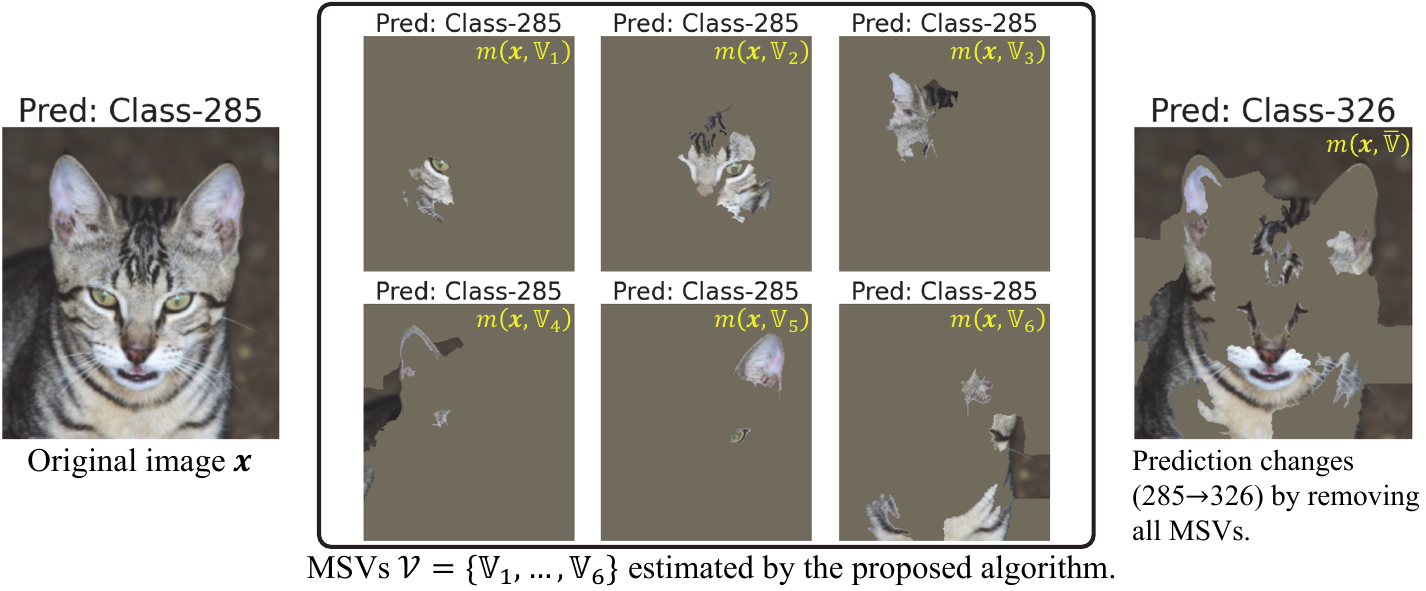}
    \caption{Example of estimated MSVs for an image in the Open Images validation set. Section~\ref{sec:msv} presents the definitions of mathematical expressions.}
    \label{fig:ex-msv}
\end{figure}

Figure~\ref{fig:ex-msv} shows an example of MSVs obtained using the algorithm described in Section~\ref{sec:greedy}.
The left side of Figure~\ref{fig:ex-msv} shows this image\footnote{
    To clarify the licenses, we used images from the Open Images dataset to illustrate the proposed method and other methods that use specific images.
    Please refer to \ref{apdx:image_urls} for URLs of the original images.
} from a validation split of the Open Images dataset~\citep{kuznetsova2020open}.
We applied a ResNet-101 model~\citep{he2016deep} trained with the ImageNet dataset~\citep{deng2009imagenet} to the image and obtained a predicted class of 285 (\textit{Egyptian Cat}) out of 1000 classes of ImageNet.
The center of Figure~\ref{fig:ex-msv} shows the MSVs estimated using the proposed algorithm with respect to the ResNet-101 prediction, which comprised six images with different masked patterns.
All these masked images were still classified as Class~285 by ResNet, indicating that there were six views in the prediction of the image, each of which was sufficient to make the prediction that the image was an \textit{Egyptian Cat}.
We can observe from these MSVs that the prediction of the left image was supported by several features, including the right eye and face stripe pattern, left eye and face stripe pattern, right ear, left ear, and body stripe pattern.
The right side of Figure~\ref{fig:ex-msv} shows an image with all MSVs inverted, resulting in a prediction change from Class 285 to Class 326 (\textit{Lycaenid}).

\subsection{Minimal sufficient views} \label{sec:msv}
We now formally define the proposed MSVs.
For simplicity, the following definitions are used for vector inputs. However, the same definitions can be applied to images and other inputs:
Given a $K$-class classification problem, let the input be~$\bm{x} \in \mathbb{R}^n$ and the prediction model be~$f:\mathbb{R}^n \rightarrow \mathbb{R}^K$.
The $i$th element of the vector $\bm{x}$ is denoted by $x_i$ and the $j$th output of the multivalued function~$f$ is denoted by $f_j$.
Let~$c_f(\bm{x}):=\argmax_{k \in \llbracket K \rrbracket} f_k(\bm{x})$ be a function that determines the prediction class for~$\bm{x}$ based on~$f$, where we use the notation$\llbracket a \rrbracket := \left\{1,\ldots,a\right\}$ for positive integer~$a$.

We denote the \emph{view} of an input~$\bm{x}$ as $\mathbb{V} \subseteq \llbracket n \rrbracket \ (\mathbb{V} \neq \emptyset)$,
which is, a subset of the indices of the elements of~$\bm{x}$.
The masked input~$m(\bm{x}, \mathbb{V})$ for an input~$\bm{x}$ and view~$\mathbb{V}$ are defined as follows:
\begin{align*}
     & m(\bm{x}, \mathbb{V}) := \left(m_1, \ldots, m_n\right)^\top,                  \quad
    m_i = \left\{ \begin{array}{ll}
                      x_i & \text{if } i \in \mathbb{V}, \\
                      b_i & \text{otherwise,}
                  \end{array} \text{ for } i \in \llbracket n \rrbracket, \right.
\end{align*}
where~$\bm{b} \in \mathbb{R}^n$ is the baseline input, which is also used in existing XAI methods, including integrated gradients~\citep{sundararajan2017axiomatic} or DeepLift\footnote{It is referred to as ``reference input'' in DeepLift.}~\citep{shrikumar2017learning}).
There is no single method for determining the baseline input and there are various options, as discussed in~\cite{sturmfels2020visualizing}.
In this study, we used the average value of~$\bm{x}$ in the training data as~$\bm{b}$ unless otherwise noted.
A comparison of the different baseline inputs used in our experiment is discussed in Section~\ref{sec:accessing_baseline}.
Examples of masked inputs are shown in the middle of Figure~\ref{fig:ex-msv}, where the gray area corresponds to the baseline~$\bm{b}$.

\begin{dfn} \label{def:min_suf}
    Given~$f$, a view $\mathbb{V}$ for~$\bm{x}$ is \emph{minimally sufficient} if
    \begin{align*}
         & \text{(Sufficiency)} \quad c_f(\bm{x}) = c_f\left(m(\bm{x}, \mathbb{V})\right),                                                              \\
         & \text{(Minimality)}  \quad c_f(\bm{x}) \neq c_f\left(m(\bm{x}, \mathbb{V}\setminus \left\{i\right\})\right), \quad \forall i \in \mathbb{V}.
    \end{align*}
\end{dfn}
By definition, sufficiency implies that the prediction of~$f$ for~$\bm{x}$ remains unchanged by applying the view, whereas minimality implies that the prediction changes after removing an element from the view.
We now give our definition of MSVs:
\begin{dfn}  \label{def:MSVs}
    Given~$f$, the \emph{MSVs}~$\mathcal{V}$ for~$\bm{x}$ is expressed as $\mathcal{V} := \left\{\mathbb{V}_1, \ldots, \mathbb{V}_q \right\}$ such that
    \begin{itemize}
        \item $\mathbb{V}_i$ is minimal sufficient $\forall i \in \llbracket q \rrbracket$,
        \item $\mathbb{V}_i \cap \mathbb{V}_j = \emptyset$ if $i \neq j$,
        \item and $c_f(\bm{x}) \neq c_f\left(m(\bm{x}, \bar{\mathbb{V}})\right)$, where $\textstyle{\bar{\mathbb{V}}=\llbracket n \rrbracket \setminus \bigcup_{i \in \llbracket q \rrbracket} \mathbb{V}_i}$.
    \end{itemize}
\end{dfn}
By definition, MSVs is a set of disjointed views, each of which is minimal sufficient, as shown in the middle of Figure~\ref{fig:ex-msv}.
The last condition indicates that~$\bar{\mathbb{V}}$ is not sufficient; that is, excluding all views in the MSVs changes the prediction, as shown on the right-hand side of Figure~\ref{fig:ex-msv}.
Note that MSVs are not necessarily unique for given $f$ and $\bm{x}$.

\subsection{Greedy algorithm} \label{sec:greedy}
It is computationally expensive to determine the exact MSVs, particularly when the input dimension $n$ is large.
Thus, we proposed a greedy algorithm to heuristically compute MSVs, wherein the minimality condition in Definition~\ref{def:min_suf} was relaxed to $\beta$-split-minimality.

\subsubsection{$\beta$-Split-Minimality}
First we introduce the \textproc{Split} function that splits a view~$\mathbb{V}$, which is an index set, into $\beta$ groups using~$\bm{x}$ as an auxiliary.
If~$\bm{x}$ is an image, superpixel methods such as SLIC~\citep{achanta2012slic} can be used as \textproc{Split}, and the Voronoi partition~\citep{de2000computational} can also be used as an $\bm{x}$-independent splitting method.
We can accelerate the search for MSVs by exploring the group units that are obtained by \textproc{Split} instead of exploring each element of~$\bm{x}$.
Note that \textproc{Split} will split~$\mathbb{V}$ into~$\beta' = \left|\mathbb{V}\right|$ groups if~$\left|\mathbb{V}\right| < \beta$.
An example of applying~\textproc{Split} is presented in Section~\ref{sec:execution_example}.

We define the $\beta$-split minimality as follows:
\begin{dfn} \label{def:beta_min}
    Given~$f$ and a \textproc{Split} function, a view $\mathbb{V}$ for~$\bm{x}$ satisfies $\beta$-split-minimality if it satisfies the following:
    \begin{align*}
         & \text{($\beta$-Split-Minimality)} \quad c_f(\bm{x}) \neq c_f\left(m(\bm{x}, \mathbb{V} \setminus \mathbb{S})\right), \quad \forall \mathbb{S} \in \text{\textproc{Split}}\left(\mathbb{V}, \bm{x}, \beta\right)
    \end{align*}
\end{dfn}
By setting~$\beta$ as sufficiently large ($\beta=\left|\mathbb{V}\right|$), the $\beta$-split-minimality becomes equivalent to the minimality condition in Definition~\ref{def:min_suf}; however, the larger~$\beta$ is, the higher the computational cost of the following \textproc{GreedyMSVs}.
Our experiments showed that a relatively small~$\beta$ (4–32) was sufficient to obtain results corresponding to Figure~\ref{fig:val_acc_msv}.

\subsubsection{GreedyMSVs}
We proposed \textproc{GreedyMSVs} in Algorithm~\ref{alg:greedy_msv} to compute MSVs that satisfied the $\beta$-split-minimality.
In Algorithm~\ref{alg:greedy_msv}, the submodule \textproc{EstimateMSV} was called recursively, and each time it was called, the view at that point was divided into $\beta$ groups by \textproc{Split}.
Thus, in the early stages of a search, when the view size is large, the search can be performed using a coarse partition.
As the search progressed and the view size decreased, the search could be performed using a more detailed partition.
An example execution is presented in the next section.
Every view in~$\mathcal{V}$ obtained by Algorithm~\ref{alg:greedy_msv} satisfied both sufficiency and $\beta$-split minimality.
\begin{algorithm}[tb]
    \caption{Greedy algorithm to compute MSVs}
    \label{alg:greedy_msv}
    \begin{small}
        \textbf{[Input]} $\bm{x}$: input to the model, $f$: classification model\\
        \textbf{[Parameter]} $\beta$: number of splits\\
        \textbf{[Output]} $\mathcal{V}$: MSVs
        \begin{algorithmic}[1]
            \Function{GreedyMSVs}{$\bm{x}$, $f$}
            \State{$\mathcal{V} \leftarrow \emptyset$, $\mathbb{V}_0 \leftarrow \llbracket n \rrbracket$}  \Comment{Initialize}
            \State{$k \leftarrow c_f\left(\bm{x}\right)$} \Comment{Predicted class for $\bm{x}$}
            \Do
            \State{$\mathbb{V} \leftarrow$ \textproc{EstimateMSV}($\bm{x}$, $f$, $\mathbb{V}_0$, $k$)} \Comment{Estimate an MSV}
            \State{$\mathcal{V} \leftarrow \mathcal{V} \cup \left\{\mathbb{V}\right\}$} \Comment{Add to $\mathcal{V}$}
            \State{$\mathbb{V}_0 \leftarrow \mathbb{V}_0 \setminus \mathbb{V}$}  \Comment{Remove used indexes}
            \DoWhile{$c_f\left(m(\bm{x}, \mathbb{V}_0)\right) = k$} \Comment{Continue as long as~$\mathbb{V}_0$ is sufficient} \label{alg:sufficient1}
            \State{\Return{$\mathcal{V}$}}
            \EndFunction
            \Function {EstimateMSV}{$\bm{x}$, $f$, $\mathbb{V}$, $k$}
            \State{$\mathcal{S} \leftarrow \textproc{Split}(\mathbb{V}, \bm{x}, \beta)$} \Comment{Split indexes into groups} \label{alg:split}
            \State{$\mathbb{S}' \leftarrow \argmin_{\mathbb{S} \in \mathcal{S}} \left|f_{k}\left(\bm{x}\right) - f_{k}\left(m(\bm{x}, \mathbb{V} \setminus \mathbb{S})\right) \right|$} \Comment{Select $\mathbb{S}'$ with minimal change} \label{alg:select_S}
            \State{$\mathbb{V}' \leftarrow \mathbb{V} \setminus \mathbb{S}'$}
            \If{$c_f\left(m(\bm{x}, \mathbb{V}')\right) = k$ and $\left|\mathbb{V}'\right|>0$} \Comment{Judge whether $\mathbb{V}'$ is sufficient or not} \label{alg:sufficient2}
            \State{$\mathbb{V} \leftarrow$ \textproc{EstimateMSV}($\bm{x}$, $f$, $\mathbb{V}'$, $k$)} \Comment{Recursively shrink $\mathbb{V}$}
            \EndIf
            \State{\Return{$\mathbb{V}$}}
            \EndFunction
        \end{algorithmic}
    \end{small}
\end{algorithm}

\subsubsection{Execution example} \label{sec:execution_example}
\begin{figure}[tb]
    \centering
    \includegraphics[width=1.0\linewidth,clip]{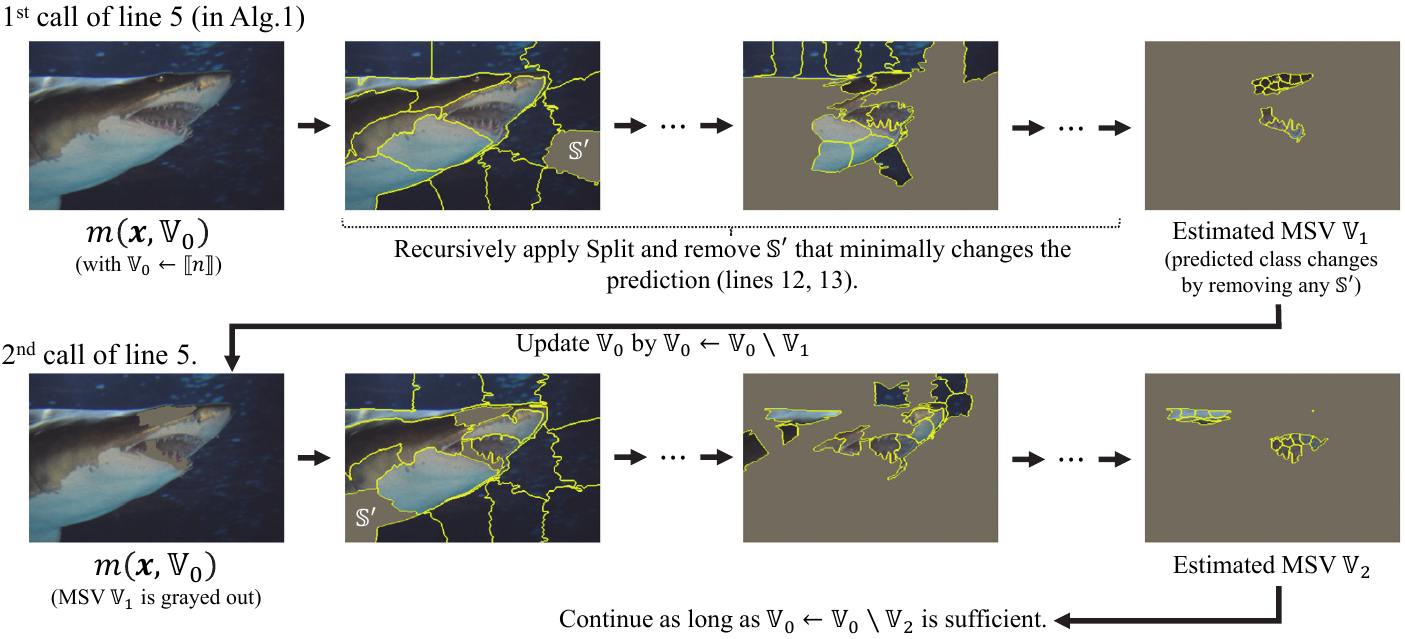}
    \caption{Execution example of MSVs search by \textproc{GreedyMSVs}.}
    \label{fig:execution}
\end{figure}

Figure~\ref{fig:execution} shows an example of searching for MSVs with \textproc{GreedyMSVs}.
We used an image from the Open Images validation set, and the ResNet-101 model was trained using ImageNet.
We used SLIC implemented in~\citep{scikit-image} as \textproc{Split}.
The yellow line in the figure shows the partition obtained by \textproc{Split} with $\beta=16$.
The upper left image shows the original image classified by the model as \textit{Tiger Shark}. Further, the next image to the right shows the result by \textproc{Split} and~$\mathbb{S}'$ was the region selected in Line~\ref{alg:select_S} of Algorithm~\ref{alg:greedy_msv}, which minimally changed the prediction.
Then~$\mathbb{S}'$ was removed from the current view and \textproc{Split} was applied recursively.
Finally, we obtained the first MSV~$\mathbb{V}_1$,
as shown in the upper-right panel of the figure,
where the prediction changed if we removed any region obtained by \textproc{Split}.
As evident from the figure, the recursive application of \textproc{Split} resulted in a gradual refinement of the resulting partition, facilitating a fine-grained adaptive search of a view while speeding up the search.
Then, MSV~$\mathbb{V}_1$ was removed from~$\mathbb{V}_0$ and the search continued as long as the updated~$\mathbb{V}_0$ was sufficient; that is, ~$m\left(\bm{x}, \mathbb{V}_0\right)$ was classified as a \textit{Tiger Shark}.

\section{Experimental results} \label{sec:experiments}

In this section, we present examples of MSVs estimation using benchmark datasets and demonstrate their relation to the generalization performance of DNNs.

\subsection{Setup for classification models}
We obtained pretrained classification models from TorchVision\footnote{\url{https://pytorch.org/vision/stable/models.html} (accessed August 2, 2023)}, listed in the legend of Figure~\ref{fig:val_acc_msv}, trained on the ImageNet training set, and used them for evaluation.
To evaluate MSVs on these DNN models, we used the ImageNet~\citep{deng2009imagenet}, Open Images~\citep{kuznetsova2020open}, ImageNet-C~\citep{hendrycks2018benchmarking}, and CIFAR-100~\citep{krizhevsky2009learning} datasets.
Unless otherwise stated, we used SLIC~\citep{achanta2012slic} implemented in~\citep{scikit-image} as \textproc{Split} in Algorithm~\ref{alg:greedy_msv} with~$\beta=16$ to compute MSVs.

\subsection{Comparison of MSVs for a single DNN model}
\subsubsection{MSVs for images of the same prediction class}
\begin{figure}[tb]
    \centering
    \includegraphics[width=1.0\linewidth,clip]{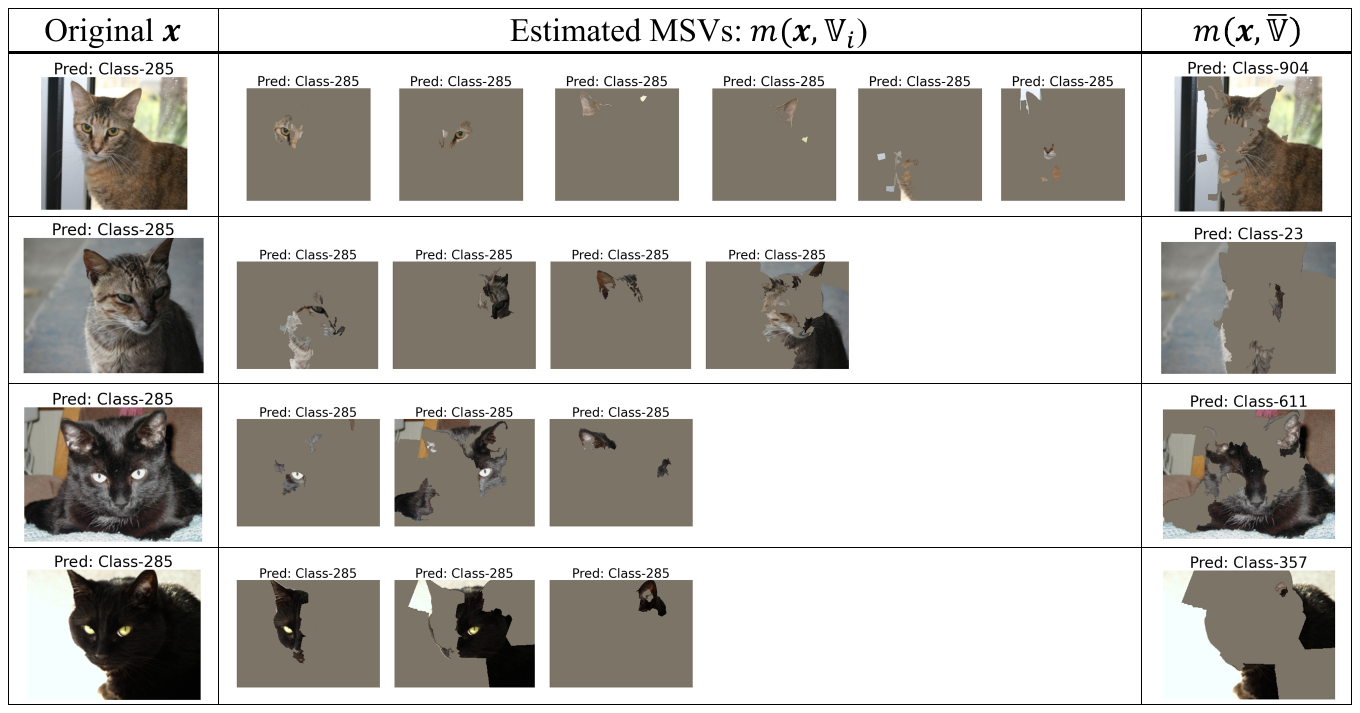}
    \caption{MSVs for images from the Open Images validation set that are predicted to be Class~285.}
    \label{fig:class285_msvs}
\end{figure}

First, we present the MSVs estimated with respect to the prediction using ResNet-101, demonstrating that the proposed algorithm extracted consistent evidence from multiple input images of the same prediction class.
We computed MSVs for images from the Open Images validation dataset, in addition to the example shown in Figure~\ref{fig:ex-msv}, each of which was predicted to be Class~285 (\textit{Egyptian Cat}) by ResNet-101.
The results are presented in Figure~\ref{fig:class285_msvs}.
Interestingly, although the MSVs were obtained separately for each image, MSVs with common features were obtained from multiple images, such as the left and right eyes and the left and right ears. This indicated that ResNet-101 used these features as evidence to classify the image as Class~285.
This result was similar to the multi-view assumption in~\citet{allen2020towards}, which assumed that there were common latent features across images and DNNs used these features to make predictions.
In \ref{apdx:msv_examples}, we provide examples of MSVs estimated from the images of other prediction classes, showing that these MSVs shared common features across images of the same prediction class.
However, there was no clear trend in the change in prediction after removing the estimated MSVs (please refer to the prediction class for $m(\bm{x}, \bar{\mathbb{V}})$ in the figure).
The number of estimated MSVs varied depending on the image, ranging as 3–6 in Figure~\ref{fig:ex-msv} and ~\ref{fig:class285_msvs}.

\subsubsection{Comparing images by estimated number of MSVs}
\begin{figure}[tb]
    \centering
    \includegraphics[width=1.0\linewidth,clip]{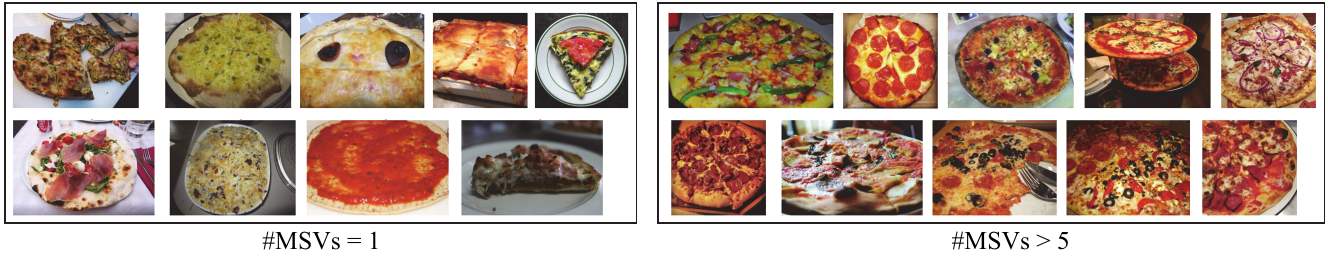}
    \caption{Images from the Open Images validation set predicted to be Class~963. The number of estimated MSVs is 1 for the left group and greater than 5 for the right group. The estimated MSVs for each image are detailed in \ref{apdx:msv_examples}.}
    \label{fig:class963_compare_n_msv}
\end{figure}

To compare the appearances of images with different numbers of estimated MSVs, we computed MSVs for images from the Open Images dataset predicted by ResNet-101 as Class~963 (\textit{Pizza}).
The left side of Figure~\ref{fig:class963_compare_n_msv} shows images with~$|\mathcal{V}|=1$, where~$\mathcal{V}$ is the estimated MSVs with respect to the prediction of Class~963, whereas the right side of the figure shows images with~$|\mathcal{V}|>5$.
The right group included images of a typical pizza with tomato sauce and cheese with toppings such as salami or olives.
In contrast, the left group contained pizza images without tomato sauce, toppings, or even non-pizza images.
These results suggested that typical pizza images comprised multiple features, each of which could be evidence that the model classified the image as a pizza, and that the proposed algorithm successfully estimated multiple MSVs for these images.
From another perspective, these results indicated that the number of estimated MSVs was related to the prediction difficulty.
In the next section, we show that the prediction accuracy is clearly related to the number of estimated MSVs.

\subsubsection{Prediction accuracy vs. the number of MSVs}
To investigate the correlation between the number of estimated MSVs and prediction accuracy of a single model,
we classified each image in the ImageNet validation set of 50000 images using ResNet-101, and computed the MSVs for each prediction.
Note that MSVs were calculated based on the predicted classes; therefore, information about the true class label was \emph{not} used to calculate MSVs.
We grouped the validation images according to the number of estimated MSVs and evaluated the prediction accuracy along with its 95~\% confidence interval for each group.\footnote{The confidence interval is computed as $\pm 1.96 \times \sqrt{p(1-p)/n}$, following \citep{casella2024statistical}, where~$n$ is the number of data points in the group and~$p$ is the accuracy within the group.}
The results are shown in Table~\ref{tab:resnet101_acc_nmsv}.
As shown in Table~\ref{tab:resnet101_acc_nmsv}, there was a clear relationship between the number of estimated MSVs and the prediction accuracy, suggesting that a DNN model provided a more reliable output if its prediction was based on more evidence estimated as MSVs.
\begin{table}[tb]
    \center
    \tabcolsep = 6pt
    \caption{Prediction accuracy of ResNet-101 for the ImageNet validation set, wherein images are grouped based on the number of estimated MSVs. Numbers in parentheses represent 95~\% confidence intervals of the prediction accuracy.}
    \label{tab:resnet101_acc_nmsv}
    \scalebox{0.65}[0.65]{
        \begin{tabular}{c|cccccccccc}
            \toprule
            \#MSVs                    & 1                            & 2                            & 3                            & 4                            & 5                            & 6                            & 7                            & 8                            & 9                            & $\geq 10$                    \\
            \midrule
            \vspace{-3pt}
            \multirow{2}{*}{Accuracy} & 0.538                        & 0.808                        & 0.865                        & 0.891                        & 0.905                        & 0.919                        & 0.919                        & 0.922                        & 0.923                        & 0.937                        \\
                                      & \footnotesize{$(\pm 0.008)$} & \footnotesize{$(\pm 0.008)$} & \footnotesize{$(\pm 0.008)$} & \footnotesize{$(\pm 0.009)$} & \footnotesize{$(\pm 0.010)$} & \footnotesize{$(\pm 0.011)$} & \footnotesize{$(\pm 0.014)$} & \footnotesize{$(\pm 0.015)$} & \footnotesize{$(\pm 0.017)$} & \footnotesize{$(\pm 0.007)$} \\
            \bottomrule
        \end{tabular}
    }
\end{table}

\subsection{Comparison of MSVs between different DNN models} \label{sec:model_comparison}
In the previous section, we demonstrated that the number of MSVs has a clear relationship with the prediction accuracy in a single DNN.
In the following, we show that such a relationship also holds between different DNNs, including the convolution and transformer models.

\subsubsection{Average number of estimated MSVs and prediction accuracy}
\begin{figure}[t]
    \centering
    \includegraphics[width=1.0\linewidth,clip]{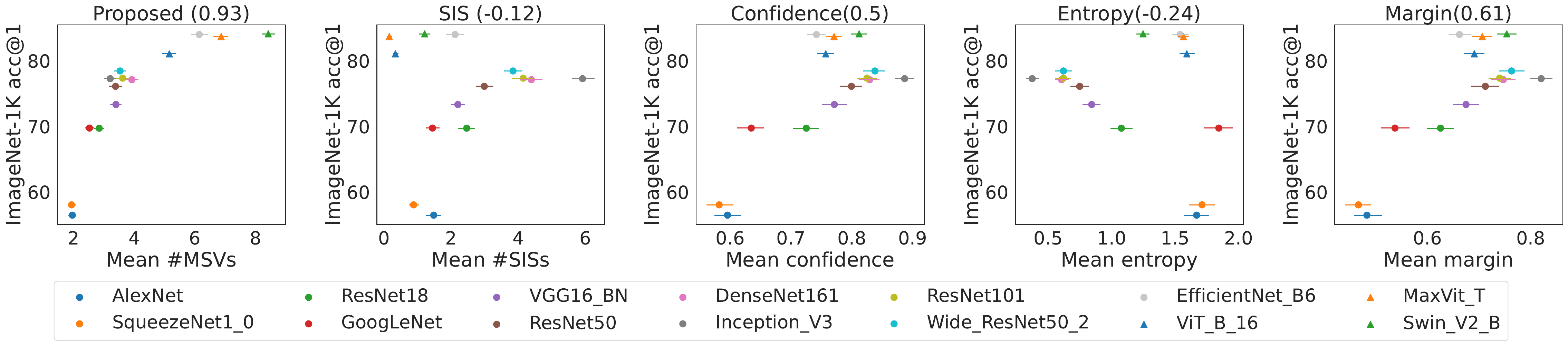}
    \caption{Average score versus top 1 accuracy. Values in parentheses for each method are rank correlations between score and accuracy.}
    \label{fig:comparison_acc_score}
\end{figure}

To investigate the relationship between the number of MSVs and the generalizability of different DNN models, we computed MSVs for various ImageNet-trained models obtained from TorchVision (listed in the legend of Figure~\ref{fig:comparison_acc_score}).
We randomly sampled 1000 images from the ImageNet validation dataset and computed MSVs for these images using each model.
The leftmost figure in Figure~\ref{fig:comparison_acc_score}, which is also shown in Figure~\ref{fig:val_acc_msv}, shows the relationship between the average number of MSVs for these 1000 images (x-axis) and the prediction accuracy over the entire validation set (y-axis) of each model.
A clear relationship in that a model with a larger number of MSVs achieved a higher prediction accuracy was observed.
Note that the numbers in parentheses in the figure indicate a rank correlation between the score on the x-axis and accuracy on the y-axis, which is as high as 0.93 for the proposed MSVs.
Thus, models that make predictions based on more evidence, estimated as MSVs, have a higher generalization performance on average.

In Figure~\ref{fig:comparison_acc_score}, the confidence intervals for the mean values are shown as lines, representing the 1\%-99\% percentile range of the mean values obtained using the bootstrap method~\citep{efron1994introduction}.
The difference in the mean number of MSVs across models typically exceeds the width of these confidence intervals. This suggests that model evaluation based on the mean number of MSVs remains reliable even when using a subset of the validation dataset.
To examine the variation in average score estimation with smaller sample sizes, we evaluated the confidence intervals of the estimated average scores by sub-sampling from the original scores obtained from 1000 images in the ImageNet validation dataset. The sub-sampling sizes were set to 100 or 10.
Figure~\ref{fig:proposed_conventional_100_10} in \ref{apdx:small_sample_size} presents the results, suggesting that a sample size of~100 still provides a reasonable estimate of the average score compared to the difference in mean scores between models. In contrast, when the sample size is set to~10, the confidence interval becomes much larger, making the average score unsuitable for model comparison.

\paragraph{Application to DNN model selection}
The leftmost figure in Figure~\ref{fig:comparison_acc_score} suggests that the average number of MSVs can be used as a metric for selecting a DNN model with better generalization performance.
As the calculation of MSVs does not require class labels, we can evaluate the metric using an unlabeled dataset.
In addition, the average number of MSVs computed using the training dataset was highly correlated with that computed using the validation dataset.
Therefore, we can use the dataset used to train the model for selection, as discussed in Section~\ref{sec:eval_with_train}.

\paragraph{Comparison with other metrics} \label{sec:comparison_with_other_metrics}
To compare the proposed metric, that is, the average number of MSVs, with existing metrics, we evaluated various metrics using the same 1000 images from the ImageNet validation set for each model, as shown in Figure~\ref{fig:comparison_acc_score}.
As an existing metric, we used the average confidence, which is shown in~\citep{9710388} as being a good predictor of the prediction accuracy of DNNs.
We also evaluated similar metrics: average entropy and average 1-vs-2 margin~\cite{settles2009active}.
We also computed SIS~\cite{carter2019made} for the same 1000 images and evaluated the average number of SIS per DNN model, where we applied BG-SIS~\citep{carter2021overinterpretation} to compute SIS approximately for computational efficiency\footnote{We used the authors' implementation for BG-SIS (\url{https://github.com/gifford-lab/overinterpretation}, accessed May 27, 2024). Further, unless otherwise noted, we used the default parameters specified in~\citep{carter2021overinterpretation}: confidence threshold was set to 0.9 and $k=100$.\label{footnote:bg-sis}}.
Note that these metrics, like the proposed one, can be evaluated without the need for label information.
The results are summarized in Figure~\ref{fig:comparison_acc_score}, where the x-axis represents the average of each metric over 1000 images, and the rank correlations between the score and accuracy are shown in parentheses.
As evident, all the metrics had lower correlations than the proposed method.
The average confidence, margin, and number of SIS showed a high correlation with accuracy when only convolutional models were considered, except for EfficientNet\_B6. However, this correlation decreased when transformer models, represented by triangles in the figure, were included. In contrast, the average number of MSVs maintained a consistent relationship across both types of models.

\subsubsection{MSVs estimation using training dataset} \label{sec:eval_with_train}
\begin{figure}[tbp]
    \centering
    \includegraphics[width=\linewidth,clip]{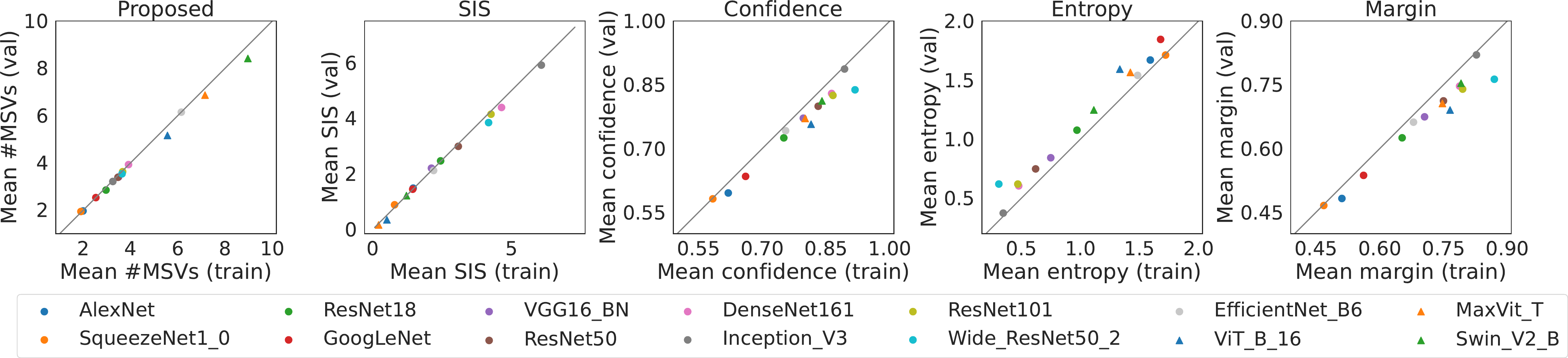}
    \caption{Comparison of scores that are estimated using the training set (x-axis) and the validation set (y-axis).}
    \label{fig:score_train_val}
\end{figure}

Instead of using images from the ImageNet \emph{validation} set, we evaluated each metric in Figure~\ref{fig:comparison_acc_score} using 1000 images that were randomly sampled from the ImageNet \emph{training} set.
Figure~\ref{fig:score_train_val} shows a comparison between the estimated metrics, where the x-axis is the metric estimated using the training set, and the y-axis is the metric estimated using the validation set.
We observed that there were gaps between the metrics estimated using the training and validation sets in the average confidence, entropy, and margin.
This is because DNN models tend to be overconfident in the data used for training.
In contrast, the gap between the metrics estimated using the training and validation sets was considerably smaller for the average number of MSVs or SIS. This suggested that the number of pieces of evidence, in terms of MSVs or SIS, used in the prediction was less dependent on the degree of overfitting of the model.
From Figure~\ref{fig:comparison_acc_score} and ~\ref{fig:score_train_val}, we can use the dataset used for training to evaluate the average number of MSVs for comparing models.
This property is extremely useful because it eliminates the need to prepare holdout data when training and selecting models.

\subsubsection{Evaluation with other classification datasets}
\begin{figure}[tbp]
    \centering  \includegraphics[width=1.0\linewidth,clip]{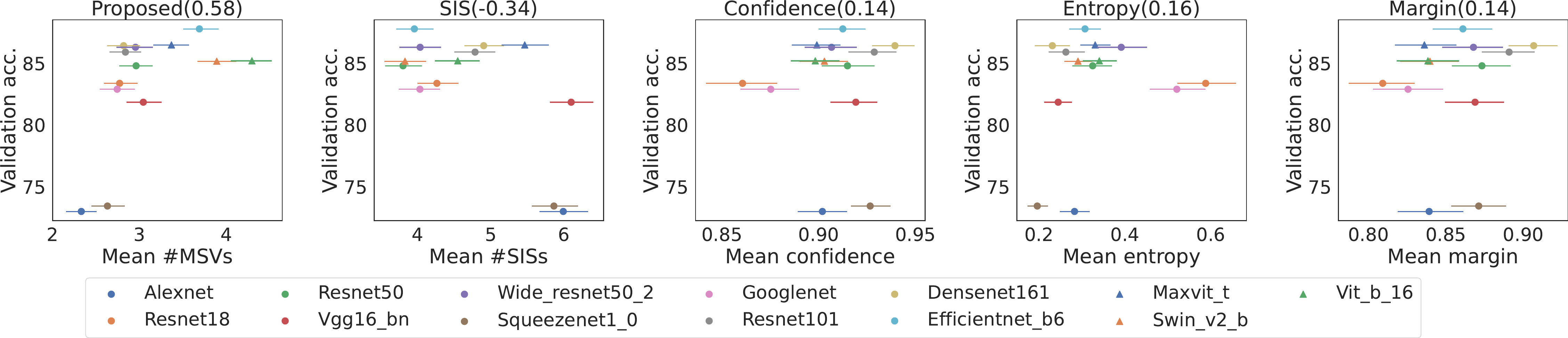}
    \caption{Average score versus top 1 accuracy for CIFAR-100 dataset. Values in parentheses for each method are rank correlations between score and accuracy.}
    \label{fig:cifar100}
\end{figure}

\paragraph{CIFAR-100 dataset}
We finetuned TorchVision's ImageNet-trained models using the CIFAR-100~\citep{krizhevsky2009learning} training set with a stochastic gradient descent for $200$~epochs and a batchsize of~$128$. Further, the weight decay was set to $0.0005$ and the initial learning rate was set to $0.01$, which was multiplied by~$0.1$ at epochs~$100$ and~$150$.
For each model, we selected the best model within the training epochs according to the accuracy in case of the CIFAR-100 validation set.
Subsequently, we computed the average number of MSVs and other metrics for these models using 1000 images randomly sampled from the CIFAR-100 validation set.
Figure~\ref{fig:cifar100} shows the results, where the x-axis is the average of each score over 1000 images, and the y-axis is the prediction accuracy over the entire CIFAR-100 validation set, with rank correlations in parentheses.
The lines in Fig.~\ref{fig:cifar100} correspond to the 1\%-99\% confidence intervals obtained through the bootstrap method.
We can observe that the average number of MSVs exhibited a higher correlation with accuracy than the other scores.

\begin{figure}[tbp]
    \centering
    \includegraphics[width=1.0\linewidth,clip]{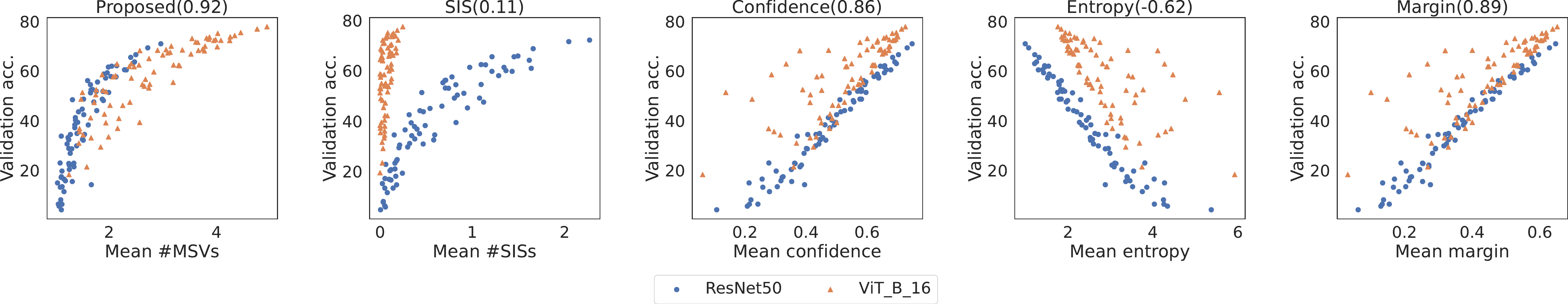}
    \caption{Average score versus top 1 accuracy for ImageNet-C dataset. Values in parentheses for each method are rank correlations between score and accuracy.}
    \label{fig:imagenet-c}
\end{figure}

\paragraph{ImageNet-C dataset}
We also evaluated the proposed metric and other metrics for the ImageNet-C dataset~\citep{hendrycks2018benchmarking}, which comprised 75 datasets created by distorting images in ImageNet with different types (15 types) and levels (five levels) of distortion.
In the experiment with ImageNet-C, we used ResNet-50 and ViT\_B\_16 as representatives of convolutional and transformer-based models, respectively, to reduce the computational burden.
The results are shown in Figure~\ref{fig:imagenet-c}, where we observe that the rank correlation between the average number of estimated MSVs and prediction accuracy remains as high as 0.92 in the ImageNet-C evaluation.
As the results suggest, increasing image distortion further reduces the number of MSVs, corresponding to a decrease in accuracy due to the distortion.

\subsection{Assessing the impact of hyperparameters in MSVs}
There are several hyperparameters in Algorithm~\ref{alg:greedy_msv} for estimating MSVs, including~$\beta$, the baseline value~$\bm{b}$, and the choice of \textproc{Split} function.
Next, we examine the impact of these hyperparameters on the estimation of MSVs.

\subsubsection{Assessing the effect of $\beta$}
\begin{figure}[tbp]
    \centering
    \includegraphics[width=\linewidth,clip]{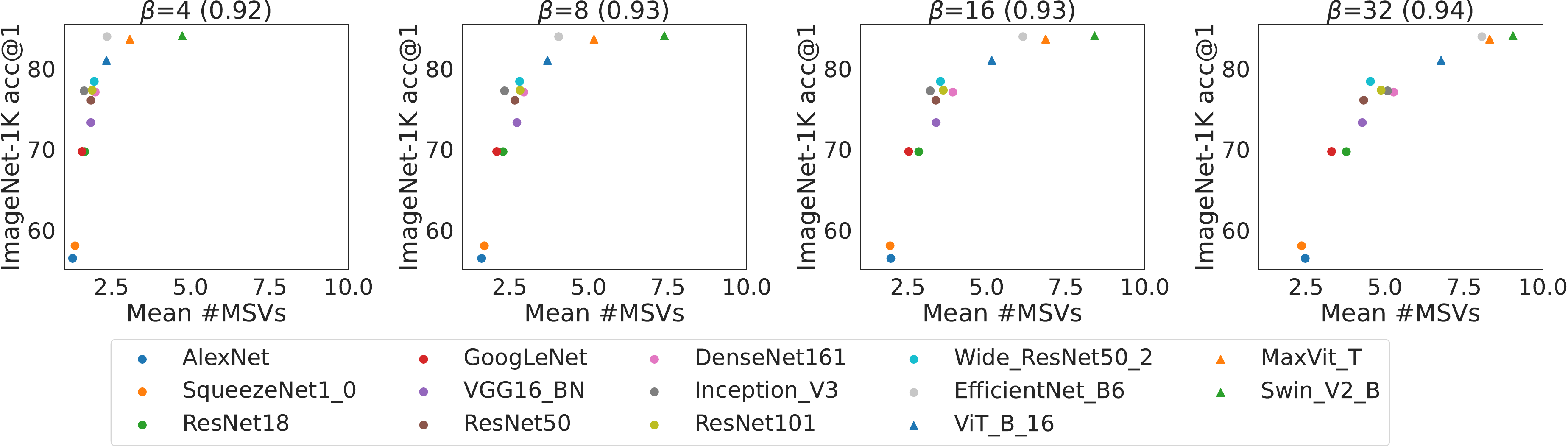}
    \caption{Average number of MSVs versus top 1 accuracy with $\beta$ changing as 4–32. Values in parentheses are rank correlations.}
    \label{fig:varying_beta}
\end{figure}

The proposed algorithm for estimating MSVs has a parameter~$\beta$, which was~$\beta=16$ in previous sections, that controls the number of groups obtained by \textproc{Split}.
To investigate the impact of~$\beta$, we computed MSVs using Algorithm~\ref{alg:greedy_msv} with~$\beta$ varying as 4–32 for randomly sampled 1000 images from the ImageNet validation set. The other experimental settings were similar to those in the previous section.
The results are shown in Figure~\ref{fig:varying_beta} with rank correlations in parentheses.
Figure~\ref{fig:varying_beta} shows that the average number of MSVs exhibited a high correlation with the accuracy, regardless of~$\beta$.
The average number of MSVs decreased as~$\beta$ decreased.
This is natural because the $\beta$-split-minimality became harder to hold as~$\beta$ decreased, leading to a smaller number of estimated MSVs.

\subsubsection{Assessing the effect of baseline value} \label{sec:accessing_baseline}
In addition to the results in Table~\ref{tab:resnet101_acc_nmsv}, we conducted experiments wherein the baseline value~$\bm{b}$ used for masking the image was white, black, or random (with a normal distribution).
Using ResNet-101, we computed MSVs using each baseline for all the images in the ImageNet validation set.
The validation images were grouped according to the number of estimated MSVs and the
prediction accuracy was evaluated in each group. The results are shown in Table~\ref{tab:baseline_value}.
The table indicates that the accuracy of images with few MSVs (one or two) was relatively low, regardless of the baseline used for masking. Accuracy consistently increased as the number of MSVs grew, particularly when using the average value as the baseline, especially for images with a large number of MSVs.
\begin{table}[tbp]
    \center
    \tabcolsep = 6pt
    \caption{Prediction accuracy for the ImageNet validation set, wherein the images are grouped based on
        the number of estimated MSVs. Numbers in parentheses represent 95~\% confidence intervals of the prediction accuracy.}
    \label{tab:baseline_value}
    \scalebox{0.65}[0.65]{
        \begin{tabular}{l|cccccccccc}
            \toprule
                                     & \#MSVs                       &                              &                              &                              &                              &                              &                              &                              &                              &                              \\
            \cmidrule{2-11}
            Baseline~$\bm{b}$        & 1                            & 2                            & 3                            & 4                            & 5                            & 6                            & 7                            & 8                            & 9                            & $\geq 10$                    \\
            \midrule
            \vspace{-3pt}
            \multirow{2}{*}{Average} & 0.538                        & 0.808                        & 0.865                        & 0.891                        & 0.905                        & 0.919                        & 0.919                        & 0.922                        & 0.923                        & 0.937                        \\
                                     & \footnotesize{($\pm 0.008$)} & \footnotesize{($\pm 0.008$)} & \footnotesize{($\pm 0.008$)} & \footnotesize{($\pm 0.009$)} & \footnotesize{($\pm 0.010$)} & \footnotesize{($\pm 0.011$)} & \footnotesize{($\pm 0.014$)} & \footnotesize{($\pm 0.015$)} & \footnotesize{($\pm 0.017$)} & \footnotesize{($\pm 0.007$)} \\
            \vspace{-3pt}
            \multirow{2}{*}{White}   & 0.595                        & 0.857                        & 0.897                        & 0.912                        & 0.915                        & 0.927                        & 0.926                        & 0.917                        & 0.931                        & 0.919                        \\
                                     & \footnotesize{($\pm 0.007$)} & \footnotesize{($\pm 0.006$)} & \footnotesize{($\pm 0.007$)} & \footnotesize{($\pm 0.009$)} & \footnotesize{($\pm 0.012$)} & \footnotesize{($\pm 0.014$)} & \footnotesize{($\pm 0.018$)} & \footnotesize{($\pm 0.023$)} & \footnotesize{($\pm 0.025$)} & \footnotesize{($\pm 0.013$)} \\
            \vspace{-3pt}
            \multirow{2}{*}{Black}   & 0.620                        & 0.854                        & 0.885                        & 0.897                        & 0.898                        & 0.911                        & 0.913                        & 0.903                        & 0.919                        & 0.914                        \\
                                     & \footnotesize{($\pm 0.007$)} & \footnotesize{($\pm 0.006$)} & \footnotesize{($\pm 0.008$)} & \footnotesize{($\pm 0.010$)} & \footnotesize{($\pm 0.013$)} & \footnotesize{($\pm 0.015$)} & \footnotesize{($\pm 0.018$)} & \footnotesize{($\pm 0.022$)} & \footnotesize{($\pm 0.024$)} & \footnotesize{($\pm 0.010$)} \\
            \vspace{-3pt}
            \multirow{2}{*}{Random}  & 0.604                        & 0.860                        & 0.898                        & 0.916                        & 0.923                        & 0.914                        & 0.922                        & 0.942                        & 0.900                        & 0.919                        \\
                                     & \footnotesize{($\pm 0.007$)} & \footnotesize{($\pm 0.006$)} & \footnotesize{($\pm 0.007$)} & \footnotesize{($\pm 0.009$)} & \footnotesize{($\pm 0.011$)} & \footnotesize{($\pm 0.016$)} & \footnotesize{($\pm 0.019$)} & \footnotesize{($\pm 0.020$)} & \footnotesize{($\pm 0.032$)} & \footnotesize{($\pm 0.015$)} \\
            \bottomrule
        \end{tabular}
    }
\end{table}

\subsubsection{Assessing the effect of splitting method}
Since SLIC uses k-means clustering in its procedure, the splitting results may change depending on the random seed used in SLIC.
To assess the robustness of our results with respect to SLIC's non-deterministic nature, we conducted experiments following the same procedure as in Table~\ref{tab:resnet101_acc_nmsv} but with different random seeds in SLIC.
Table~\ref{tab:slic_diff_seed} presents the results, suggesting that the relationship between accuracy and the number of MSVs remains stable regardless of the seed.
\begin{table}[tbp]
    \center
    \tabcolsep = 6pt
    \caption{Prediction accuracy for the ImageNet validation set, wherein images are grouped based on
        the number of estimated MSVs. ``Seed'' refers to the random seed used in SLIC (results with $\mathrm{Seed}=0$  correspond to Table~\ref{tab:resnet101_acc_nmsv}).}
    \label{tab:slic_diff_seed}
    \scalebox{0.65}[0.65]{
        \begin{tabular}{l|cccccccccc}
            \toprule
                               & \#MSVs                       &                              &                              &                              &                              &                              &                              &                              &                              &                              \\
            \cmidrule{2-11}
            Seed               & 1                            & 2                            & 3                            & 4                            & 5                            & 6                            & 7                            & 8                            & 9                            & $\geq 10$                    \\
            \midrule
            \multirow{2}{*}{0} & 0.538                        & 0.808                        & 0.865                        & 0.891                        & 0.905                        & 0.919                        & 0.919                        & 0.922                        & 0.923                        & 0.937                        \\
                               & \footnotesize{$(\pm 0.008)$} & \footnotesize{$(\pm 0.008)$} & \footnotesize{$(\pm 0.008)$} & \footnotesize{$(\pm 0.009)$} & \footnotesize{$(\pm 0.010)$} & \footnotesize{$(\pm 0.011)$} & \footnotesize{$(\pm 0.014)$} & \footnotesize{$(\pm 0.015)$} & \footnotesize{$(\pm 0.017)$} & \footnotesize{$(\pm 0.007)$} \\
            \multirow{2}{*}{1} & 0.539                        & 0.800                        & 0.871                        & 0.892                        & 0.905                        & 0.908                        & 0.921                        & 0.935                        & 0.935                        & 0.932                        \\
                               & \footnotesize{($\pm 0.008$)} & \footnotesize{($\pm 0.008$)} & \footnotesize{($\pm 0.008$)} & \footnotesize{($\pm 0.009$)} & \footnotesize{($\pm 0.010$)} & \footnotesize{($\pm 0.012$)} & \footnotesize{($\pm 0.013$)} & \footnotesize{($\pm 0.014$)} & \footnotesize{($\pm 0.016$)} & \footnotesize{($\pm 0.007$)} \\
            \multirow{2}{*}{2} & 0.538                        & 0.803                        & 0.861                        & 0.897                        & 0.903                        & 0.913                        & 0.933                        & 0.918                        & 0.940                        & 0.936                        \\
                               & \footnotesize{($\pm 0.008$)} & \footnotesize{($\pm 0.008$)} & \footnotesize{($\pm 0.008$)} & \footnotesize{($\pm 0.009$)} & \footnotesize{($\pm 0.010$)} & \footnotesize{($\pm 0.012$)} & \footnotesize{($\pm 0.012$)} & \footnotesize{($\pm 0.016$)} & \footnotesize{($\pm 0.016$)} & \footnotesize{($\pm 0.007$)} \\
            \multirow{2}{*}{3} & 0.535                        & 0.804                        & 0.871                        & 0.893                        & 0.908                        & 0.914                        & 0.907                        & 0.923                        & 0.932                        & 0.938                        \\
                               & \footnotesize{($\pm 0.008$)} & \footnotesize{($\pm 0.008$)} & \footnotesize{($\pm 0.008$)} & \footnotesize{($\pm 0.009$)} & \footnotesize{($\pm 0.010$)} & \footnotesize{($\pm 0.011$)} & \footnotesize{($\pm 0.014$)} & \footnotesize{($\pm 0.015$)} & \footnotesize{($\pm 0.017$)} & \footnotesize{($\pm 0.007$)} \\
            \multirow{2}{*}{4} & 0.537                        & 0.803                        & 0.869                        & 0.891                        & 0.906                        & 0.919                        & 0.914                        & 0.926                        & 0.939                        & 0.934                        \\
                               & \footnotesize{($\pm 0.008$)} & \footnotesize{($\pm 0.008$)} & \footnotesize{($\pm 0.008$)} & \footnotesize{($\pm 0.009$)} & \footnotesize{($\pm 0.010$)} & \footnotesize{($\pm 0.011$)} & \footnotesize{($\pm 0.013$)} & \footnotesize{($\pm 0.015$)} & \footnotesize{($\pm 0.015$)} & \footnotesize{($\pm 0.007$)} \\

            \bottomrule
        \end{tabular}
    }
\end{table}

To further investigate the impact of the splitting method on the proposed method, we used the Voronoi partition~\citep{de2000computational} as a splitting method in addition to SLIC with~$\beta=8$ or~$\beta=16$.
The results are listed in Table~\ref{tab:compare_split}, where the ResNet-101 and the ImageNet validation sets were used.
From the table, it is evident that there was a clear relationship between the accuracy and number of estimated MSVs, regardless of the splitting method used in Algorithm~\ref{alg:greedy_msv}.
\begin{table}[tbp]
    \center
    \tabcolsep = 6pt
    \caption{Prediction accuracy for the ImageNet validation set, wherein images are grouped based on
        the number of estimated MSVs.}
    \label{tab:compare_split}
    \scalebox{0.63}[0.63]{
        \begin{tabular}{l|cccccccccc}
            \toprule
                                                  & \#MSVs                       &                              &                              &                              &                              &                              &                              &                              &                              &                              \\
            \cmidrule{2-11}
            \textproc{Split} function             & 1                            & 2                            & 3                            & 4                            & 5                            & 6                            & 7                            & 8                            & 9                            & $\geq 10$                    \\
            \midrule

            \vspace{-3pt}
            \multirow{2}{*}{SLIC ($\beta=16$)}    & 0.538                        & 0.808                        & 0.865                        & 0.891                        & 0.905                        & 0.919                        & 0.919                        & 0.922                        & 0.923                        & 0.937                        \\
                                                  & \footnotesize{($\pm 0.008$)} & \footnotesize{($\pm 0.008$)} & \footnotesize{($\pm 0.008$)} & \footnotesize{($\pm 0.009$)} & \footnotesize{($\pm 0.010$)} & \footnotesize{($\pm 0.011$)} & \footnotesize{($\pm 0.014$)} & \footnotesize{($\pm 0.015$)} & \footnotesize{($\pm 0.017$)} & \footnotesize{($\pm 0.007$)} \\
            \vspace{-3pt}
            \multirow{2}{*}{SLIC ($\beta=8$)}     & 0.571                        & 0.843                        & 0.899                        & 0.918                        & 0.928                        & 0.933                        & 0.944                        & 0.938                        & 0.965                        & 0.947                        \\
                                                  & \footnotesize{($\pm 0.007$)} & \footnotesize{($\pm 0.007$)} & \footnotesize{($\pm 0.007$)} & \footnotesize{($\pm 0.008$)} & \footnotesize{($\pm 0.010$)} & \footnotesize{($\pm 0.012$)} & \footnotesize{($\pm 0.015$)} & \footnotesize{($\pm 0.019$)} & \footnotesize{($\pm 0.017$)} & \footnotesize{($\pm 0.011$)} \\
            \vspace{-3pt}
            \multirow{2}{*}{Voronoi ($\beta=16$)} & 0.531                        & 0.785                        & 0.841                        & 0.871                        & 0.895                        & 0.885                        & 0.902                        & 0.908                        & 0.909                        & 0.927                        \\
                                                  & \footnotesize{($\pm 0.008$)} & \footnotesize{($\pm 0.009$)} & \footnotesize{($\pm 0.010$)} & \footnotesize{($\pm 0.010$)} & \footnotesize{($\pm 0.011$)} & \footnotesize{($\pm 0.013$)} & \footnotesize{($\pm 0.014$)} & \footnotesize{($\pm 0.015$)} & \footnotesize{($\pm 0.016$)} & \footnotesize{($\pm 0.005$)} \\
            \vspace{-3pt}
            \multirow{2}{*}{Voronoi ($\beta=8$)}  & 0.586                        & 0.848                        & 0.892                        & 0.909                        & 0.921                        & 0.925                        & 0.931                        & 0.929                        & 0.931                        & 0.948                        \\
                                                  & \footnotesize{($\pm 0.007$)} & \footnotesize{($\pm 0.007$)} & \footnotesize{($\pm 0.007$)} & \footnotesize{($\pm 0.009$)} & \footnotesize{($\pm 0.011$)} & \footnotesize{($\pm 0.013$)} & \footnotesize{($\pm 0.016$)} & \footnotesize{($\pm 0.019$)} & \footnotesize{($\pm 0.021$)} & \footnotesize{($\pm 0.009$)} \\

            \bottomrule
        \end{tabular}
    }
\end{table}

\begin{figure}[htbp]
    \centering
    \includegraphics[width=1.0\linewidth,clip]{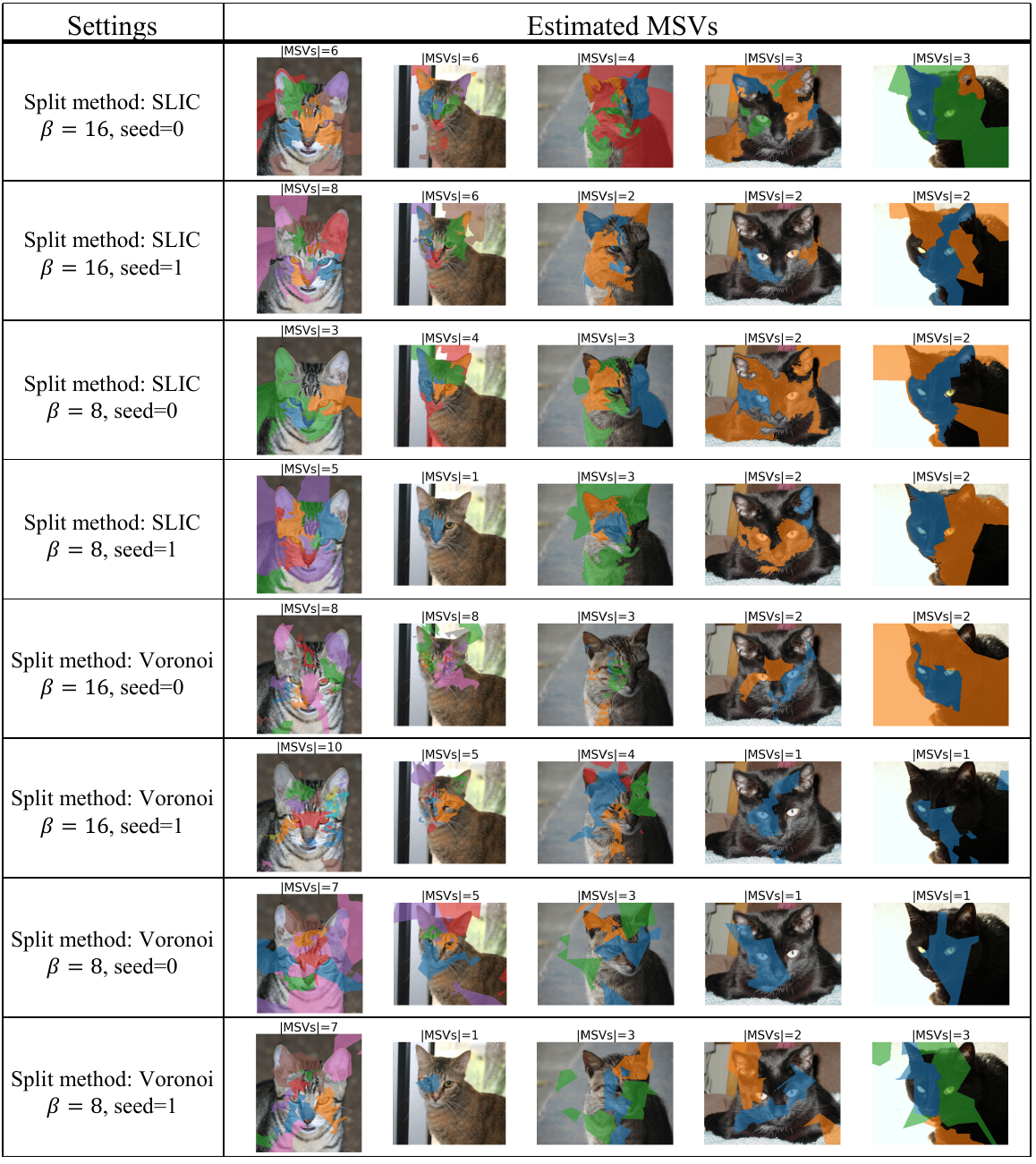}
    \caption{Comparison of estimated MSVs using different splitting methods,~$\beta$, and random seeds. Each colored region represents an estimated MSV.}
    \label{fig:compare_split_beta}
\end{figure}

Next, in Figure~\ref{fig:compare_split_beta}, we compare the MSVs obtained under different settings of the \textproc{Split} function using the cat images from Figures~\ref{fig:execution} and~\ref{fig:class285_msvs}.
The estimated MSVs are presented in a single image, with each colored region representing a single MSV. For example, the MSVs in Figure~\ref{fig:ex-msv} are summarized in the upper-left image of Figure~\ref{fig:compare_split_beta}.
In Figure~\ref{fig:compare_split_beta}, we present results for eight different combinations of SLIC or Voronoi, $\beta = 8$ or $\beta = 16$, and a random seed of 0 or 1.
The first observation from Figure~\ref{fig:compare_split_beta} was that the MSVs changed depending on the splitting method, the value of~$\beta$, and the random seed.
This was primarily because the MSVs satisfying Definition~\ref{def:MSVs} were not uniquely determined, as discussed in Section~\ref{sec:msv}.
The second key observation from Figure~\ref{fig:compare_split_beta} was that the number of estimated MSVs generally follows a consistent order, except for a few cases. The estimated number of MSVs is displayed in the upper part of each image.
These results suggested that the number of MSVs had a consistent tendency. However, the estimated MSVs for each image should be interpreted with caution because of their dependence on several hyperparameters.
A possible approach to reduce the dependence of MSVs on hyperparameters may be to estimate a common pattern from MSVs of multiple images with different hyperparameters.
An example of this approach was reported in \citep{carter2019made}, where clustering methods were used to estimate common patterns.

\subsection{Computation time}
We evaluated the computation time of Algorithm 1 with ResNet-101 using 1000 images sampled from the ImageNet validation set used in the evaluation of Figure~\ref{fig:varying_beta}.
The experiment was conducted using a single machine with a single GPU (Nvidia A100).
The results are summarized in Table~\ref{tab:resnet101_comp_time}, where the computation time was averaged over 1000 images.
As evident, the computation time decreased owing to the use of a smaller~$\beta$. Further, less than a second (on average) was required to compute the MSVs per image when~$\beta=4$. Thus, it would take less than 8 min to compute the MSVs for these 1000 images.
Because MSVs can be computed in parallel for each image, we can further accelerate the computation of MSVs by applying parallel computing.
\begin{table}[tbp]
    \center
    \tabcolsep = 6pt
    \caption{Computation time of estimating MSVs with Algorithm~\ref{alg:greedy_msv}.}
    \label{tab:resnet101_comp_time}
    \scalebox{0.9}[0.9]{
        \begin{tabular}{c|cccc}
            \toprule
            $\beta$                      & 4    & 8    & 16   & 32    \\
            \midrule
            Average time per image (sec) & 0.46 & 1.97 & 8.48 & 41.56 \\
            \bottomrule
        \end{tabular}
    }
\end{table}

\paragraph{Comparison with BG-SIS algorithm in terms of computation time}
As discussed in Section~\ref{sec:related_work}, \citet{carter2021overinterpretation} proposed the BG-SIS algorithm to compute the SIS, which is also defined by the sufficiency and minimality conditions.
In our experiment described in Section~\ref{sec:comparison_with_other_metrics}, an average of 73.7~s per image was required to estimate the SIS using the BG-SIS algorithm for 1000 images from the ImageNet validation dataset.
By contrast, the proposed algorithm could estimate MSVs considerably faster, as shown in Table~\ref{tab:resnet101_comp_time}, mainly owing to the two differences between these algorithms.
First, the proposed algorithm did not require gradient computation, whereas it required for BG-SIS, as discussed in Section~\ref{sec:related_work}.
The second was the difference in the minimality conditions between the SIS and MSVs.
The minimality in SIS was defined, following the notation used in this paper, as ``no subset of~$\mathbb{V}$ satisfies the sufficiency.'' Whereas, the minimality of MSVs was defined as ``$\mathbb{V}$ no longer satisfies the sufficiency when an element is removed from $\mathbb{V}$,'' as shown in Definition~\ref{def:min_suf}.
The difference in the minimalities resulted in a difference in the termination conditions of the algorithms. While \textproc{EstimateMSV} immediately terminated the search when sufficiency was violated, the BG-SIS algorithm did not (as described in the BackSelect procedure in~\citep{carter2021overinterpretation}).
To further accelerate the computation of MSVs, we proposed $\beta$-split-minimality as a more relaxed minimality condition, as discussed in Section~\ref{sec:greedy}.

\subsection{Comparison with existing XAI methods} \label{sec:exp_xai}
As shown in Figure~\ref{fig:ex-msv}, the proposed MSVs enabled us to determine which part of a given image was sufficient to maintain the prediction results of a given classification model.
We can use MSVs as an XAI method to investigate how a given model makes predictions.
For example, from Figure~\ref{fig:ex-msv}, we can observe that the model predicts that the image is an \textit{Egyptian Cat} from multiple perspectives, such as the right eye and face stripe pattern, left eye and face stripe pattern, right ear, left ear, and body stripe pattern.

\begin{figure}[tb]
    \centering
    \includegraphics[width=1.0\linewidth,clip]{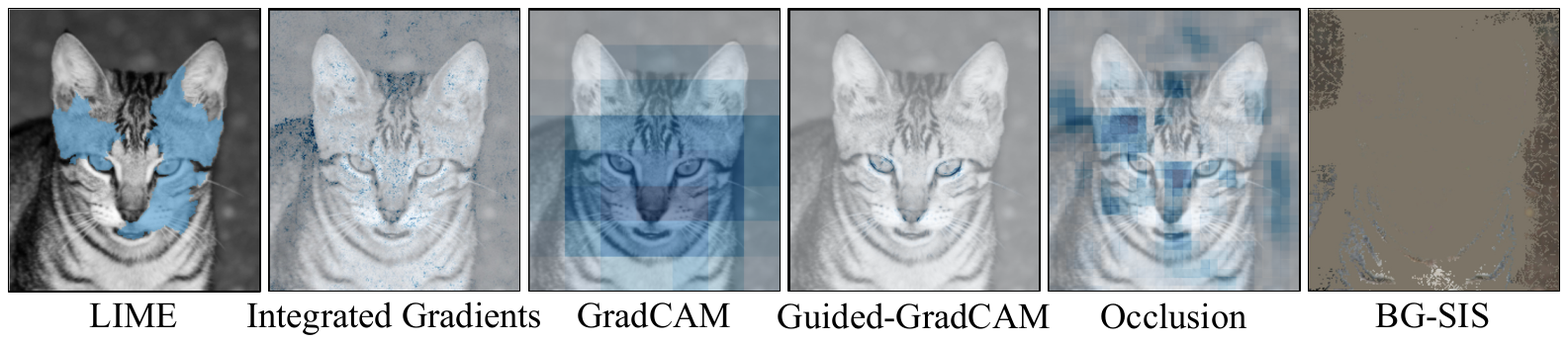}
    \caption{Results of applying other XAI methods to the prediction in Figure~\ref{fig:ex-msv}.}
    \label{fig:other_xai}
\end{figure}

To compare MSVs with other XAI methods, we applied existing XAI methods, including LIME~\citep{ribeiro2016should}, integrated gradients~\citep{sundararajan2017axiomatic}, GradCAM~\citep{selvaraju2017grad}, guided-GradCAM~\citep{selvaraju2017grad}, and occlusion ~\citep{zeiler2014visualizing}, to the prediction of the image in Figure~\ref{fig:ex-msv}.
We used the authors' implementation\footnote{\url{https://github.com/marcotcr/lime} (accessed July 27, 2023)} for LIME and Captum~\citep{kokhlikyan2020captum} for the other methods.
Figure~\ref{fig:other_xai} shows the results obtained using these methods\footnote{We used the default parameters of each library. To apply GradCAM, we used the last layer in the 4th residual layer of ResNet-101.}.
The visualization results varied depending on the method used: some highlighted the eyes, others emphasized the nose, and still others focused on the ears.
One possible reason for these inconsistent results is that the model's predictions can be explained from multiple perspectives.
Explaining the prediction of a highly nonlinear DNN using a single perspective, such as a single heat map, would be insufficient. In contrast, the proposed MSVs allow for a more comprehensive investigation of DNN predictions by explicitly considering multiple views.

We also computed the SIS for the prediction in Figure~\ref{fig:ex-msv} using the BG-SIS algorithm\footnote{We used the confidence threshold of 0.8 in BG-SIG because the default value of 0.9 resulted in no SIS being found.}.
The results are presented in Figure~\ref{fig:other_xai}.
We observed that the estimated SIS comprised mainly of the background region, which was consistent with the results reported in~\citep{carter2021overinterpretation}.
One reason for BG-SIS tending to specify the background as SIS, in contrast to the proposed algorithm, is discussed in detail in \ref{apdx:msv_sis}.
These differences between BG-SIS and the proposed algorithm are important for explaining the generalization performance of DNNs.
In fact, as shown in Section~\ref{sec:comparison_with_other_metrics}, the number of MSVs (i.e., the amount of evidence extracted around the target object) was more clearly correlated with the generalization performance of DNNs, whereas the number of SIS (i.e., the number of less interpretable patterns in the background) was not.

\subsection{Application to a detection model} \label{sec:detection}
\begin{figure}[tb]
    \centering
    \includegraphics[width=1.0\linewidth,clip]{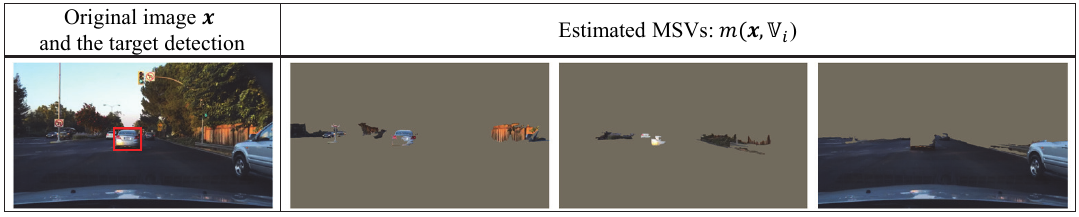}
    \caption{Estimated MSVs for the detection in the red box in the left image.}
    \label{fig:detection}
\end{figure}

We computed MSVs for a detection model using YOLOP~\citep{wu2022yolop} trained on the BDD100K dataset~\citep{yu2020bdd100k}.
We must define~$f$ appropriately to apply Algorithm ~\ref{alg:greedy_msv}.
The details for customizing~$f$ for the detection model are described in \ref{apdx:detection}.
Figure~\ref{fig:detection} shows an example of the computed MSVs for an image from the BDD100K validation set, where we computed MSVs for the detection marked by the red box in the leftmost image.
Figure~\ref{fig:detection} shows that the detection was supported by three pieces of evidence: the rear lights and rear window of the center car, part of the rear of the car and the car next to it, and the lower part of the car and road.
As demonstrated by this example, we can apply MSVs to various models by properly setting~$f$ in Algorithm~\ref{alg:greedy_msv}.

\section{Conclusion and future work} \label{sec:conclusion}
This study proposed MSVs as a set of minimal and distinct features in the input, such that each feature preserved the model's prediction for the input. Consequently, we proposed a greedy algorithm to efficiently compute approximate MSVs.
We experimentally showed that the number of MSVs corresponded to the prediction accuracy, which held true for various models, including the convolution and transformer models.
Our results suggested that DNN prediction was more reliable if the prediction was supported by more evidence estimated as MSVs.
Since the computation of MSVs does not require labels, the average number of estimated MSVs can be used as a metric for selecting a better-performing DNN model, even with an unlabeled dataset.
Further, the proposed MSVs can be used as an XAI method that can explain the prediction of DNNs from multiple perspectives.

This study can be extended in several ways.
The first can be the application of MSVs to prediction models whose inputs are not images, such as language models.
We believe that MSVs have a wide range of applications because they only require the input-output relationship of the model, that is, the model can be a black box.
The second approach can be the application of MSVs to active learning.
Because we can compute MSVs for unlabeled datasets, it is expected that we can find hard examples from unlabeled datasets by selecting images with a small number of MSVs.

\appendix

\section{Examples of estimated MSVs} \label{apdx:msv_examples}
\begin{figure}[htbp]
    \centering
    \includegraphics[width=1.0\linewidth,clip]{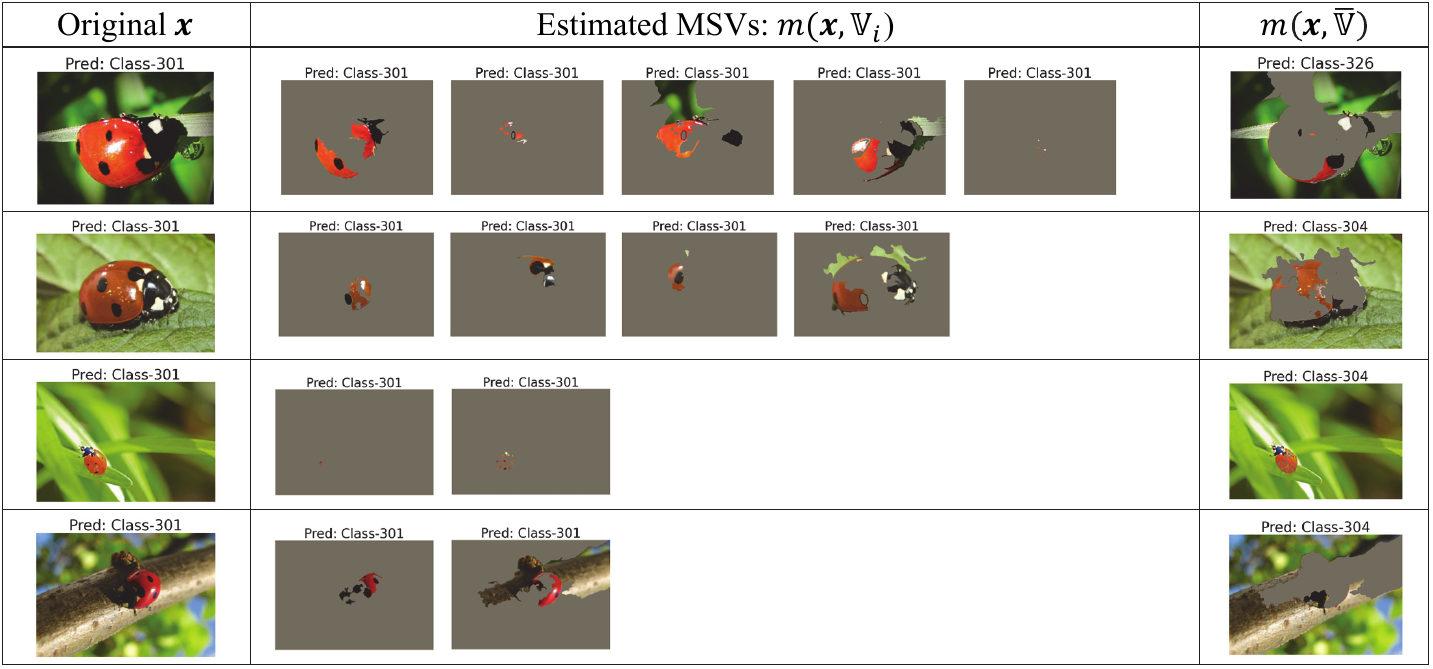}
    \caption{MSVs for images from the Open Images validation set that are predicted to be Class 301 by ResNet-101.}
    \label{fig:class301_msvs}
\end{figure}

\begin{figure}[htbp]
    \centering
    \includegraphics[width=1.0\linewidth,clip]{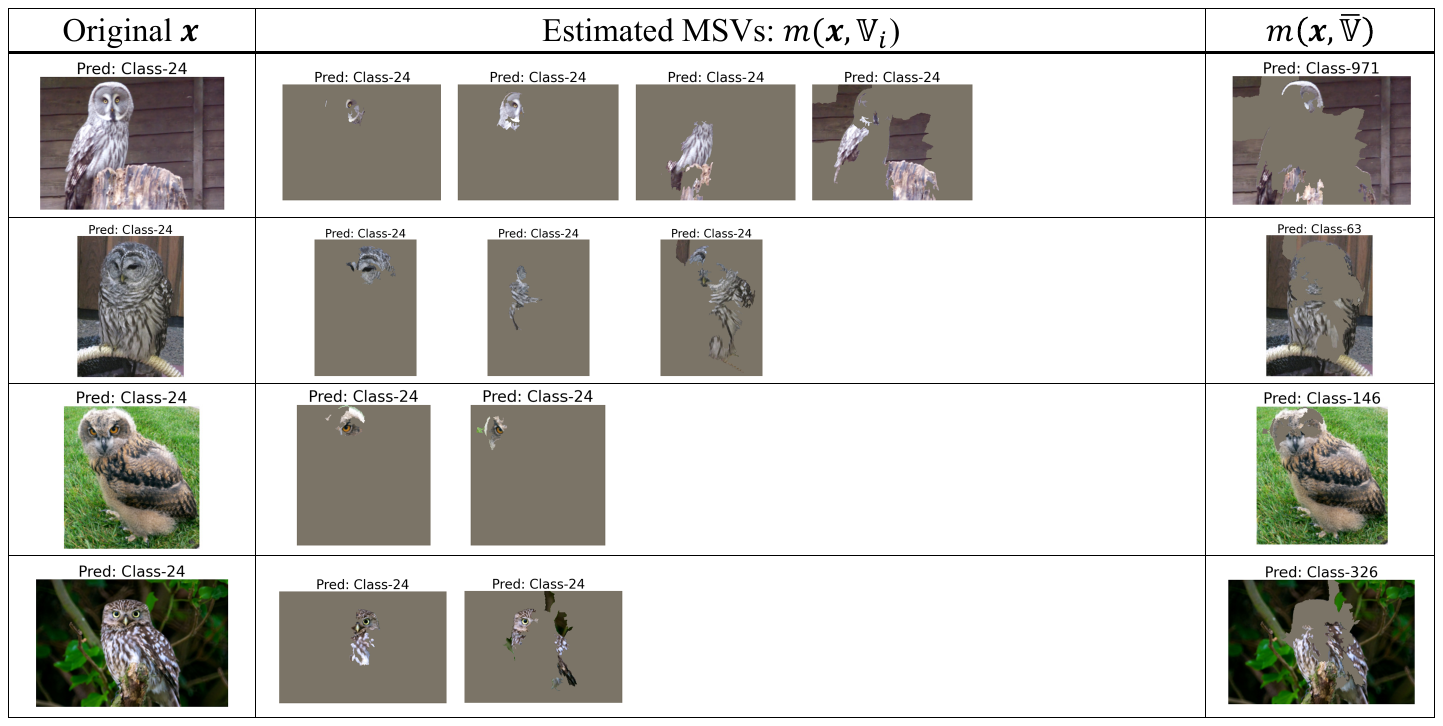}
    \caption{MSVs for images from the Open Images validation set that are predicted to be Class 24 by ResNet-101.}
    \label{fig:class24_msvs}
\end{figure}

\begin{figure}[htbp]
    \centering
    \includegraphics[width=1.0\linewidth,clip]{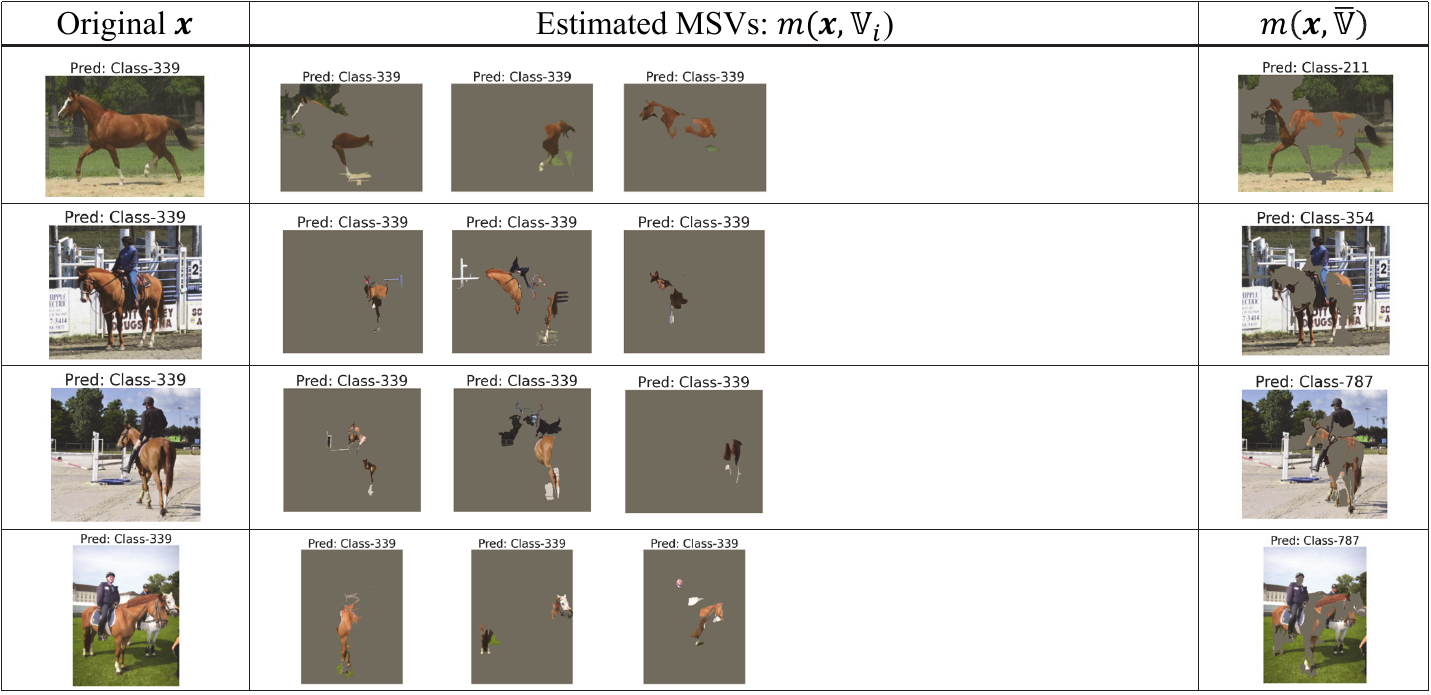}
    \caption{MSVs for images from the Open Images validation set that are predicted to be Class 339 by ResNet-101.}
    \label{fig:class339_msvs}
\end{figure}

We computed MSVs for the images from the Open Images validation dataset using TorchVision's ImageNet-trained ResNet-101.
Figures~\ref{fig:class301_msvs}–\ref{fig:class339_msvs} show the results, including the images predicted by ResNet-101 as classes 301 (\textit{Ladybug}), 24 (\textit{Great grey owl}), and 339 (\textit{Sorrel}).
MSVs with common features were obtained from multiple images that were predicted to be of the same class.
For example, the images predicted to be \textit{Ladybug} yielded MSVs containing black dots in the red pattern.
For images predicted to be the \textit{ great grey owl}, MSVs showed the right or left side of the face as a common pattern.
For images predicted to be \textit{Sorrel}, parts of the leg with muscles were estimated as shared MSVs.
Interestingly, MSVs with common features were estimated although they were estimated separately for each image, suggesting a relationship between MSVs and multi-view in~\citet{allen2020towards}.

\begin{figure}[tbhp]
    \centering
    \subfloat[Images predicted as Class~963 with $\#\mathrm{MSVs} = 1$.]{\label{fig:Class963_msvs_1}\includegraphics[width=0.9\linewidth,clip]{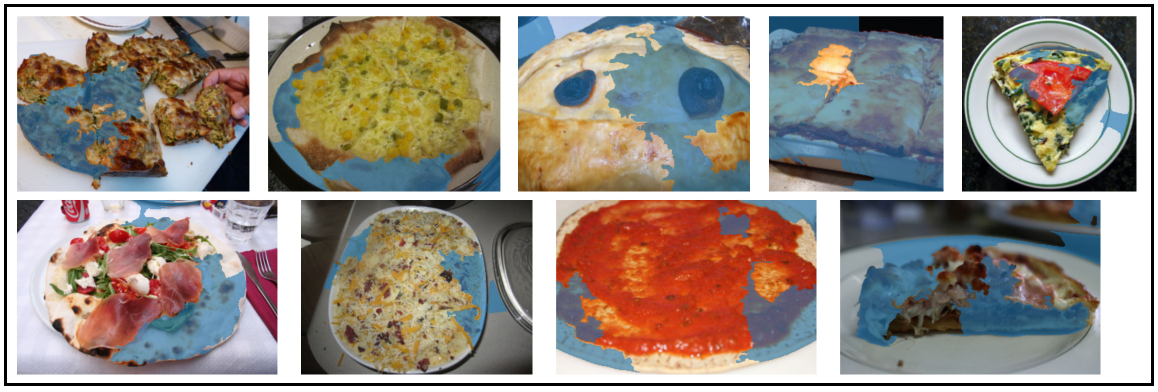}} \\
    \subfloat[Images predicted as Class~963 images with $\#\mathrm{MSVs} > 5$.]{\label{fig:Class963_msvs_5}\includegraphics[width=0.9\linewidth,clip]{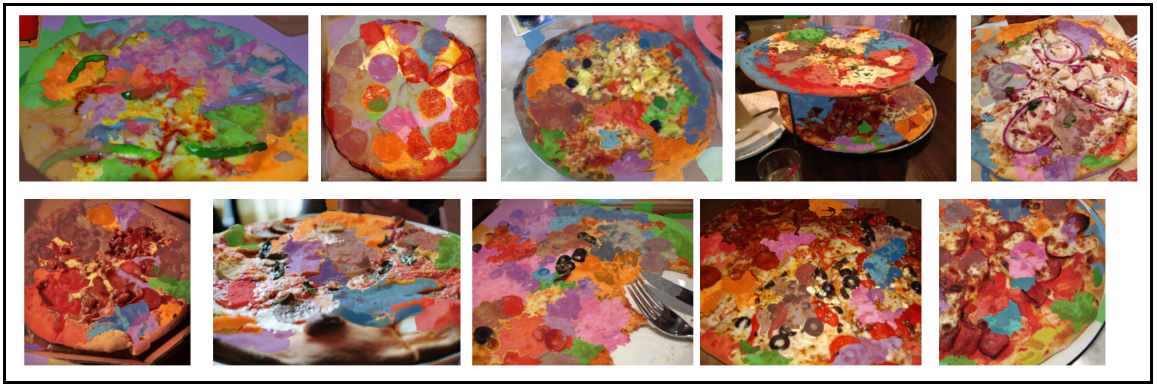}}
    \caption{Estimated MSVs for the images in Figure~\ref{fig:class963_compare_n_msv}. Each colored region represents an MSV.}
    \label{fig:Class963_msvs}
\end{figure}

In Figure~\ref{fig:Class963_msvs}, we present the MSVs estimated for the images shown in Figure~\ref{fig:class963_compare_n_msv}. Each estimated MSV is displayed within a single image, with each colored region representing a distinct MSV.

\section{Estimation of average scores with smaller sample sizes} \label{apdx:small_sample_size}
\begin{figure}[tbhp]
    \centering
    \subfloat[With a sample size of 100.]{\label{fig:proposed_conventional_100}\includegraphics[width=1.0\linewidth,clip]{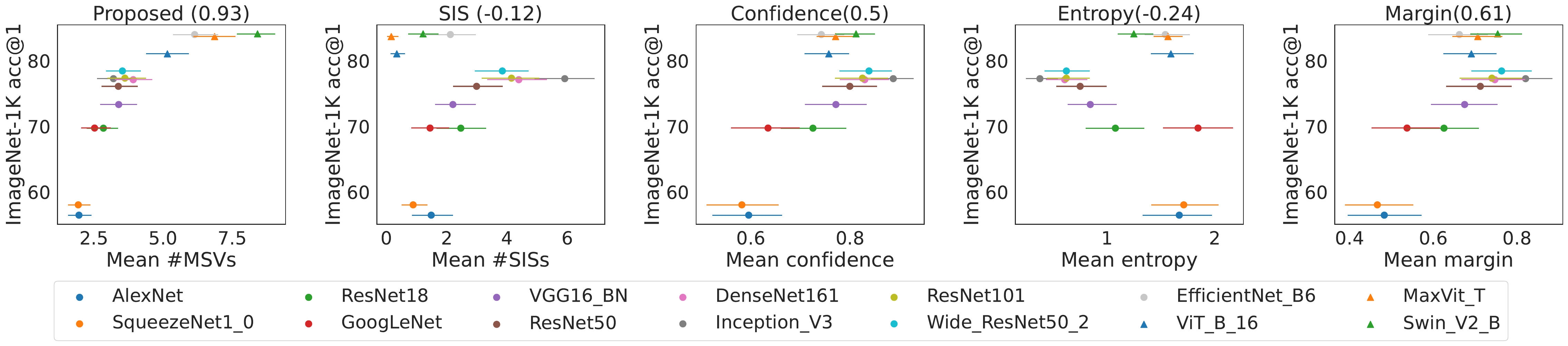}} \\
    \subfloat[With a sample size of 10.]{\label{fig:proposed_conventional_10}\includegraphics[width=1.0\linewidth,clip]{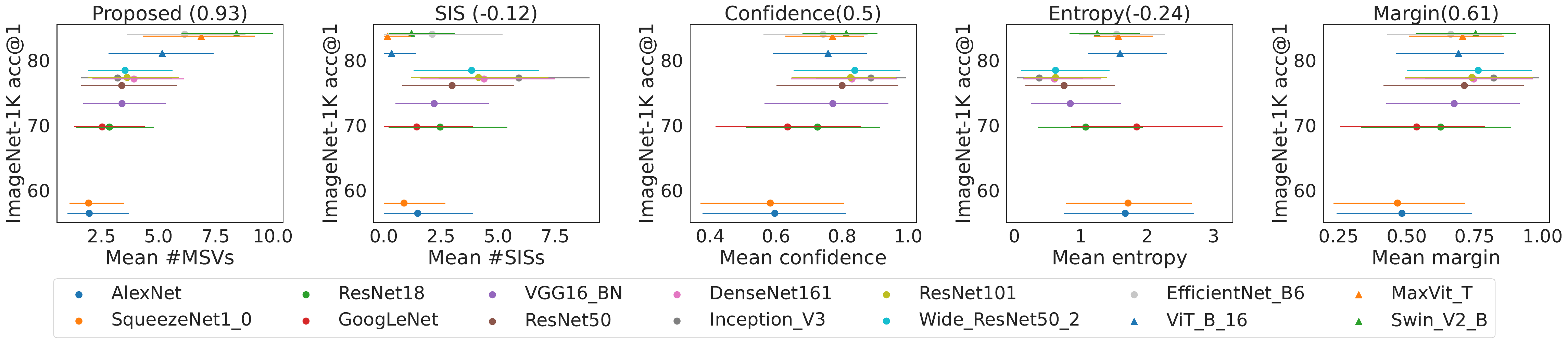}}
    \caption{Average values (points) and 1\%-99\% confidence intervals (lines) for different sample sizes.}
    \label{fig:proposed_conventional_100_10}
\end{figure}

To examine the variation in the estimation of average scores in Figure~\ref{fig:comparison_acc_score} with smaller sample sizes, we estimated each mean score by sub-sampling from the original scores obtained from 1000 images in the ImageNet validation dataset. The sub-sampling sizes were set to 100 or 10.
Figure~\ref{fig:proposed_conventional_100_10} presents the results, where the points represent the estimated mean values, and the lines indicate the 1\%-99\% confidence intervals.

\section{Algorithmic comparison of MSVs and SIS} \label{apdx:msv_sis}
The SIS~\citep{carter2019made,carter2021overinterpretation} is similar to our MSVs in that it also considers sufficiency and minimality.
However, MSVs and SIS extracted from the same image are often significantly different.
\citet{carter2021overinterpretation} reported that their BG-SIS algorithm tended to find SIS located outside the target object in an image, that is, in the background region. This was confirmed by our experiments in Section~\ref{sec:exp_xai}.
In contrast, \textproc{GreedyMSVs} in Algorithm~\ref{alg:greedy_msv} exhibited a tendency to find MSVs located around the target object that appeared to have clear semantic meanings, as shown in our experiments in Section~\ref{sec:experiments}.
This difference was mainly due to the different search strategies in BG-SIS and \textproc{GreedyMSVs}.
The BackSelect procedure in BG-SIS searches for an SIS by masking, with priority, \emph{ the part that will increase the output the most if masked} based on the gradient information.
Considering the classification for natural images by DNNs, prediction confidence can be increased by masking a small part of the image, which is the opposite of adversarial attacks ~\citep{43405}.
If such a part is more likely to be around the target object, then the BackSelect procedure in BG-SIS tends to prioritize the masking parts around the target object, resulting in an estimated SIS outside the target object.
In contrast, \textproc{GreedyMSVs} searches for an MSV by masking, with priority, \emph{a part that minimally changes the output when masked}, as described in Line~\ref{alg:select_S} of Algorithm~\ref{alg:greedy_msv}.
As such a part is more likely to be outside the target object, \textproc{GreedyMSVs} tends to mask the background part with priority, resulting in an estimated MSVs around the target object.

\section{Details on applying to a detection model}  \label{apdx:detection}
As a detection model, we used YOLOP~\citep{wu2022yolop} trained on the BDD100K dataset~\citep{yu2020bdd100k} to detect cars on the road\footnote{We used the pretrained model downloaded from \url{https://pytorch.org/hub/hustvl_yolop} (accessed July 27, 2023).}.
Given a detection model and traffic scene image~$\bm{x}$, we applied Algorithm~\ref{alg:greedy_msv} as follows:
\begin{enumerate}
    \item Detect cars in the given image using the detection model with detection threshold~$\xi \in (0,1)$, and select a detected box as the target for calculating MSVs.
    \item As YOLOP is a type of detection model that uses predefined prior boxes in its prediction, we specified the prior box to be used to detect the target.
    \item Let the detection probability of the specified prior box be~$p_{\text{det}}(\bm{x})$, we define~$f$ used in Algorithm~\ref{alg:greedy_msv} as~$f(\bm{x})=\left(p_{\text{det}}(\bm{x}), \xi\right)$. Note that Sufficiency condition used in lines~\ref{alg:sufficient1} and~\ref{alg:sufficient2} of Algorithm~\ref{alg:greedy_msv}  means that~$p_{\text{det}}(\bm{x}) \geq \xi$.
\end{enumerate}
Figure~\ref{fig:detection} in the main text shows an example of the calculated MSVs for an image from the BDD100K validation set, where we used~$\xi=0.25$ and the red box is the target detection.

\section{Image URLs} \label{apdx:image_urls}
The original credits for the images used in this study can be found on the websites listed below (accessed November 17, 2023).
All images were modified to visualize the execution results of the proposed algorithm and existing XAI methods.

\begin{itemize}
    \item Figure~\ref{fig:ex-msv}, Figure~\ref{fig:other_xai}, and Figure~\ref{fig:compare_split_beta}:
          \begin{itemize}
              \item \url{https://www.flickr.com/photos/46785529@N00/422908475}
          \end{itemize}
    \item Figure~\ref{fig:execution}:
          \begin{itemize}
              \item    \url{https://www.flickr.com/photos/yvonne_n_1968/542453357}
          \end{itemize}
    \item Figure~\ref{fig:class285_msvs} and Figure~\ref{fig:compare_split_beta}:
          \begin{itemize}
              \item \url{https://www.flickr.com/photos/24070291@N06/5760932696}
              \item \url{https://www.flickr.com/photos/ke_netan_to/97744015}
              \item \url{https://www.flickr.com/photos/mlspooky/208259954}
              \item \url{https://www.flickr.com/photos/dharmabum1964/3317040642}
          \end{itemize}
    \item Figure~\ref{fig:class963_compare_n_msv} and Figure~\ref{fig:Class963_msvs}:
          \begin{itemize}
              \item \url{https://www.flickr.com/photos/rusty_clark/9671631975}
              \item \url{https://www.flickr.com/photos/takaokun/4342613657}
              \item \url{https://www.flickr.com/photos/8143264@N08/6396168851}
              \item \url{https://www.flickr.com/photos/dana_moos/14263516186}
              \item \url{https://www.flickr.com/photos/llstalteri/7562636142}
              \item \url{https://www.flickr.com/photos/triller/2800056925}
              \item \url{https://www.flickr.com/photos/84335369@N00/3707715946}
              \item \url{https://www.flickr.com/photos/naterivers/7410981534}
              \item \url{https://www.flickr.com/photos/elsiehui/10354089956}
              \item \url{https://www.flickr.com/photos/hulagway/6047627858}
              \item \url{https://www.flickr.com/photos/nojofoto/8291625820}
              \item \url{https://www.flickr.com/photos/moe/3515442082}
              \item \url{https://www.flickr.com/photos/cogdog/8690615113}
              \item \url{https://www.flickr.com/photos/raybouk/16051918921}
              \item \url{https://www.flickr.com/photos/tomsbrain/7504267106}
              \item \url{https://www.flickr.com/photos/chunso/8437950173}
              \item \url{https://www.flickr.com/photos/dichohecho/4179438703}
              \item \url{https://www.flickr.com/photos/shoshanah/3449328732}
              \item \url{https://www.flickr.com/photos/ilove9and23/17294405089}
          \end{itemize}
    \item Figure~\ref{fig:class301_msvs}:
          \begin{itemize}
              \item \url{https://www.flickr.com/photos/hamed/154621901}
              \item \url{https://www.flickr.com/photos/xstuntkidx/4003674646}
              \item \url{https://www.flickr.com/photos/rhettmaxwell/2654140978}
              \item \url{https://www.flickr.com/photos/crisphotos/2472401181}
          \end{itemize}
    \item Figure~\ref{fig:class24_msvs}:
          \begin{itemize}
              \item \url{https://www.flickr.com/photos/tonyaustin/4008969588}
              \item \url{https://www.flickr.com/photos/picturesbyann/11115869665}
              \item \url{https://www.flickr.com/photos/jurvetson/159436192}
              \item \url{https://www.flickr.com/photos/pmunks/4663205678}
          \end{itemize}
    \item Figure~\ref{fig:class339_msvs}:
          \begin{itemize}
              \item \url{https://www.flickr.com/photos/9208231@N02/1120334380}
              \item \url{https://www.flickr.com/photos/kelleydenz/4529545564}
              \item \url{https://www.flickr.com/photos/psc49/8800179476}
              \item \url{https://www.flickr.com/photos/quinet/8560668127}
          \end{itemize}
\end{itemize}

\bibliographystyle{plainnat}
\bibliography{references}

\end{document}